\documentclass[runningheads]{llncs}
\usepackage{graphicx}
\usepackage[czech,english]{babel}

\usepackage{xcolor}
\usepackage{bm}
\usepackage{algorithm}
\usepackage{algorithmic}
\usepackage{mathtools}
\usepackage{subcaption}
\usepackage{amssymb}

\newcommand{\argminF}{\mathop{\mathrm{argmin}}\limits}

\newcommand{\dset}{\mathcal{D}}
\newcommand{\model}{\mathcal{M}}

\begin{document}
\title{Quantification of Predictive Uncertainty via Inference-Time Sampling}

%
%
 
\author{Katar\'{i}na T\'{o}thov\'{a}\inst{1}\orcidID{0000-0001-5864-179X} \** \and \v{L}ubor Ladick\'{y} \inst{1,2} \and Daniel Thul\inst{1} \and Marc Pollefeys \inst{1,3} \and Ender Konukoglu\inst{1} \**\**}

\authorrunning{K. T\'{o}thov\'{a} et al.}
%
\institute{ETH Zurich, Switzerland \\
\and Apagom AG, Zurich, Switzerland \\
\and Microsoft Mixed Reality and AI lab, Zurich, Switzerland\\
\** \email{katarina.tothova@inf.ethz.ch}
\**\** \email{ender.konukoglu@vision.ee.ethz.ch}}
\maketitle              
\begin{abstract}
Predictive variability due to data ambiguities has typically been addressed via construction of dedicated models with built-in probabilistic capabilities that are trained to predict uncertainty estimates as variables of interest.
These approaches require distinct architectural components and training mechanisms, may include restrictive assumptions and exhibit overconfidence, i.e., high confidence in imprecise predictions.
In this work, we propose a post-hoc sampling strategy for estimating predictive uncertainty accounting for data ambiguity. 
The method can generate different plausible outputs for a given input and does not assume parametric forms of predictive distributions. 
It is architecture agnostic and can be applied to any feed-forward deterministic network without changes to the architecture or training procedure. Experiments on regression tasks on imaging and non-imaging input data show the method's ability to generate diverse and multi-modal predictive distributions, and a desirable correlation of the estimated uncertainty with the prediction error.

\end{abstract}

\section{Introduction}
Estimating uncertainty in deep learning (DL) models' predictions, i.e., predictive uncertainty, is of critical importance in a wide range of applications from diagnosis to treatment planning, e.g.,~\cite{Begoli2019}. One generic formulation for predictive uncertainty, i.e., $p(y|x)$, using DL is through the following probabilistic model 
\begin{equation}\label{eqn:main}
    p(y|x) \triangleq \int_{\model}\int_{\dset}\int_{\theta} p(y|x,\theta, \dset, \model)dp(\theta|\dset,\model)dp(\dset)dp(\model),
\end{equation}
where $x,y,\theta,\model$ and $\dset$ represent input, output, model parameters, model specification, and training set, respectively. The joint distribution $p(y,\dset,\model,\theta|x)$ is modeled with the factorization $p(y|x,\theta,\dset,\model)p(\theta|\dset,\model)p(\dset)p(\model)$ and $p(y|x)$ is defined through marginalization of $\theta,\dset$ and $\model$.

Different components contribute to $p(y|x)$ in the marginalization and represent different sources of uncertainty. Borrowing terminology from~\cite{tanno}, $p(y|x,\theta,\dset,\model)$ is often considered as the \emph{aleatoric} component that models input-output ambiguities, i.e., when the input may not uniquely identify an output~\cite{WANG1996}. $p(\theta|\dset,\model)$ on the other hand is considered the \emph{epistemic} component stemming from parameter uncertainty or model bias~\cite{Draper}, which can be alleviated by training on more data or using a more appropriate model. Modeling $p(\theta)\triangleq \int_{\model}\int_{\dset} p(\theta|\model, \dset)dp(\dset)dp(\model)$ as the epistemic component is also possible, however, this is prohibitively costly in DL models, and therefore mostly not done. Here, we focus on the aleatoric component $p(y|x,\theta,\dset,\model)$ and propose a model agnostic method. 

The common approach to model $p(y|x,\theta,\dset,\model)$ in the recent DL literature is to build models that predict probability distributions instead of point-wise predictions.
Pioneering this effort, Tanno et al.~\cite{tanno_miccai,tanno} and Kendall and Gal~\cite{kendall2017} simultaneously proposed models that output pixel-wise factorized distributions for pixel-wise prediction problems, such as segmentation.
Going beyond this simplified factorization, more recent models predict distributions modeling dependencies between multiple outputs, notably for pixel-wise prediction models, such as~\cite{probunet,phiseg,tothova2018,tothova2020}.
These later models are more apt for real world applications as they are capable of producing realistic samples from the modelled $p(y|x,\theta,\dset,\model)$. 
On the downside, (i) these models require special structures in their architectures, and therefore the principles cannot be easily applied to any top performing architecture without modification, and (ii) they rely on the assumption that input-output ambiguities present in a test sample will be similarly present in the training set, so that models can be trained to predict posterior distributions. 

In this work, we propose a novel model for $p(y|x,\theta,\dset,\model)$ and the corresponding Metropolis-Hasting (MH)~\cite{MH} scheme for sampling of network outputs for a given input, which can be used during inference. 
We restrict ourselves to problems where a prior model for the outputs, i.e., $p(y)$, can be estimated. 
While this may seem limiting, on the contrary this restriction is often \emph{exploited} in medical image analysis to improve model robustness and generalization, e.g.,~\cite{tothova2020,oktay_acnn,milletari2017,karani}. 
Inspired by the ideas of network inversion via gradient descent~\cite{image_backprop} and neural style transfer~\cite{style_transfer}, our main contribution is a new definition of a \emph{likelihood} function that can be used to evaluate the MH acceptance criterion. 
The new likelihood function uses \emph{input backpropagation} and distances in the input space, which (i) makes it architecture agnostic, (ii) does not require access to training or ground truth data, (iii) avoids explicit formulation of analytically described energy functionals, or (iv) implementation of dedicated neural networks (NNs) and training procedures. 

We present experiments with two regression problems and compare our method with state-of-the-art methods~\cite{MCDropout,tothova2018,RIO}, as well as MC Dropout~\cite{MCDropout} for completeness, even though it is a method for quantifying epistemic uncertainty. Our experimental evaluation focuses on regression problems, however, the proposed technique can be easily applied to classification problems as well.

\section{Related work}
Post-hoc uncertainty quantification of trained networks has been previously addressed by Qiu et al.~\cite{RIO}. In their work (RIO), the authors augment a pre-trained deterministic network with a Gaussian process (GP) with a kernel built around the network's inputs and outputs. The GP can be used for a post-hoc improvement of the network's outputs and to assess the predictive uncertainty. This model can be applied to any standard NN without modifications. While mathematically elegant and computationally inexpensive, this approach requires access to training data and impose that the posteriors be normally distributed. 

One of the dedicated DL models addressing aleatoric uncertainty prediction with an assumption that a prior over outputs $p(y)$ is available is probPCA~\cite{tothova2018,tothova2020}. Developed for parametric surface reconstruction from images using a principal component analysis (PCA) to define $p(y)$, the method predicts multivariate Gaussian distributions over output mesh vertices, by first predicting posterior distributions over PC representation for a given sample. While the model takes into account covariance structure between vertices, it also requires specific architecture and makes a Gaussian assumption for the posterior. 

Sampling techniques built on Monte Carlo integration~\cite{monte_carlo} provide a powerful alternative. Traditional Markov Chain Monte Carlo (MCMC) techniques~\cite{Neal93probabilisticinference} can construct chains of proposed samples $y'$ with strong theoretical guarantees on their convergence to posterior distribution~\cite{gelman_bayes_book}. In MH~\cite{MH}, this is ensured by evaluation of acceptance probabilities. This involves calculation of a prior $p(y')$ and a likelihood $L(y'; x)= p(x|y',\theta,\dset,\model)$. At its most explicit form, the evaluation of $L(y'; x)$ would translate to the generation of plausible inputs for every given $y'$ and then calculation of the relative density at the input $x$. 

In some applications, the likelihood function can be defined analytically with energy functionals~\cite{energy_functional1,energy_functional2,energy_functional3}. General DL models, however, do not have convenient closed form formulations that allow analytical likelihood definitions. Defining a tractable likelihood with invertible neural networks~\cite{MintNET} or reversible transition kernels~\cite{MC_nets,song} is possible, but these approaches require specialized architectures and training procedures. To the best of our knowledge, a solution that can be applied to any pre-trained network has not yet been proposed. Sampling methods without likelihood evaluation, i.e., Approximate Bayesian Computation (ABC) methods~\cite{ABC:Rubin}, can step up to the task, however, they rely on sampling from a likelihood, hence the definition of an appropriate likelihood remains open.

\section{Method}\label{sec:MCMC:method}
We let $f(x|\theta)$ be a deep neural network that is trained on a training set of $(x,y)$ pairs and $\theta$ representing its parameters. The network is trained to predict targets from inputs, i.e., $y \approx f(x|\theta)$. We would like to asses the aleatoric uncertainty $p(y|x,\theta,\dset,\model)$ associated with $f(x|\theta)$. Note that while $f(x|\theta)$ is a deterministic mapping, if there is an input-output ambiguity around an $x$, then a trained model will likely show high sensitivity around that $x$, i.e., predictions will vary greatly with small changes in $x$. We exploit this sensitivity to model $p(y|x,\theta,\dset,\model)$ using $f(x|\theta)$. Such modeling goes beyond a simple sensitivity analysis~\cite{saltelli2004sensitivity} by allowing drawing realistic samples of $y$ from the modeled distribution.
%
\subsection{Metropolis-Hastings for Target Sampling}\label{mcmc} 
Our motivation comes from the well established MH MCMC methods for sampling from a posterior distribution~\cite{MH}.
For a given input $x$ and an initial state $y^0$, the MH algorithm generates Markov chains (MC) of states $\{y^t, t=1, 2, \dots, n\}$. 
At each step $t$ a new proposal is generated $y'\sim g(y'|y^t)$ according to a symmetric proposal distribution $g$. 
The sample is then accepted with the probability
\begin{equation}\label{eq:MH}
A(y', y^t |x) =  \min \bigg(1, \underbrace{\frac{g(y^t|y',x)}{g(y'|y^t,x)}}_{\textrm{transitions}} \underbrace{\frac{p(x|y',\theta,\dset,\model)}{p(x|y^t,\theta,\dset,\model)}}_{\textrm{likelihoods}}  \underbrace{\frac{p(y')}{p(y^t)}}_{\textrm{priors}}\bigg)
\end{equation}
and the next state is set as $y^{t+1}=y'$ if the proposal is accepted, and $y^{t+1}=y^t$ otherwise.
The sufficient condition for asymptotic convergence of the MH MC to the posterior $p(y|x,\theta,\dset,\model)$ is satisfied thanks to the reversibility of transitions $y^t\rightarrow y^{t+1}$~\cite{gelman_handbook}.
The asymptotic convergence also means that for arbitrary initialization, the initial samples are not guaranteed to come from $p(y|x)$, hence a burn-in period, where initial chain samples are removed, is often implemented~\cite{gelman_handbook}. 
The goal of the acceptance criterion is to reject the unfeasible target proposals $y'$ according to prior and likelihood probabilities.
The critical part here is that for every target proposal $y'$, the prior probability $p(y')$ and the likelihood $p(x|y')$ needs to be evaluated. 
While the former is feasible for the problems we focus on, the latter is not trivial to evaluate.

\subsection{Likelihood Evaluation}\label{sec:likelihood}
In order to model $p(y|x, \theta,\dset,\model)$, unlike prior work that used a dedicated network architecture, we define a likelihood function $p(x|y,\theta,\dset,\model)$ that can be used with any pre-trained network.
Given $f$ and a proposed target $y'$, like~\cite{MC_level_sets,MC_shapes,ertunc,neal2012bayesian}, we define likelihood with an energy function as 
\begin{equation}
    \label{eq:MCMC:our_likelihood}
    p(x|y', f(\cdot,\theta)) \propto \exp(-\beta \,E(x,y')),
\end{equation}
where $\beta$ is a "temperature" parameter and $E(x,y')$ is evaluated through a process we call gradient descent \textit{input backpropagation} inspired by neural network inversion approximation in~\cite{image_backprop} and neural style transfer work~\cite{style_transfer}. 
To evaluate $E(x,y')$, we generate an input sample $x'_{y'}$ that is as close as possible to $x$ and lead to the proposed shape $y'=f(x'_{y'}|\theta)$. This action can be formulated as an optimisation problem 
\begin{equation}\label{eq:backprop}
    x'_{y'} = \argminF_{x'} \, \lambda \underbrace{\rho(x', x)}_{\textrm{input loss}} + \underbrace{\mu(f(x'),y')}_{\textrm{target loss}},
\end{equation}
where $\rho: \mathcal{X}\times\mathcal{X} \rightarrow \mathbb{R}^{+}$, $\mu: \mathcal{Y}\times\mathcal{Y} \rightarrow \mathbb{R}^{+}$ are distances and $\lambda \in \mathbb{R}$ is a scaling constant ensuring proper optimisation of both loss elements. $\mu$ can be defined as the original distance used for training $f(x|\theta)$. We then define
\begin{equation}\label{eq:input_loss}
    E(x,y') \coloneqq \rho(x'_{y'}, x).
\end{equation}
We set $\rho(\cdot, x)$ as the squared $L_2$ distance in this work, i.e., $\rho(x'_{y'}, x) = \|x'_{y'} - x\|_2^2$, but other options are possible, as long as (\ref{eq:input_loss}) combined with (\ref{eq:MCMC:our_likelihood}) define a proper probability distribution over $x$. Note that the $L_2$ distance corresponds to a Gaussian distribution centered around $x'_{y'}$ as $p(x|y',\theta,\dset,\model)$. Even though $p(x|y',\theta,\dset,\model)$ is a Gaussian, it is crucial to note that this does not correspond to $p(y|x,\theta,\dset,\model)$ being a Gaussian due to the non-linearity the optimization in Equation~\ref{eq:backprop} introduces. Different $y'$'s can lead to very different $x'_{y'}$, and aggregating the samples through the MH MC process can lead to complex and multi-modal distributions as confirmed by the experiments in Section~\ref{sec:results}.

Within the MCMC context, the minimisation formulation ensures that we only generate $x'_{y'}$ close to the test image $x$ that also lead to the proposed $y'$ as the prediction.
Provided the two terms in (\ref{eq:backprop}) are balanced correctly, a low likelihood value assigned to a proposal $y'$ indicates $y'$ either lies outside the range of $f$ (the trained network is incapable of producing $y'$ for any input), or outside of the image (understood as a subset of codomain of $f$) of the inputs similar to $x$ under $f$.
On the other hand, a high likelihood value  means an $x'_{y'}$ very close to $x$ can produce $y'$ implying the sensitivity of the model around $x$. We can then consider $y'$ a sample from $p(y|x,\theta,\dset,\model)$.

The choice of $\beta$ parameter in (\ref{eq:MCMC:our_likelihood}) is important as it affects the acceptance rate in the MH sampling. Correctly setting $\beta$ ensures the acceptance ratio is not dominated by either $p(y)$ or $p(x|y,\theta,\dset,\model)$. 
The optimisation problem (\ref{eq:backprop}) needs to be solved for every MCMC proposal $y'$. We will refer to the full proposed MH MCMC sampling scheme as \textbf{Deep MH}. 

\subsection{Shape Sampling Using Lower Dimensional Representations}\label{sec:prior}
As mentioned, we focus on applications where $p(y)$ can be estimated. For low dimensional targets, this can be achieved by parametric or non-parametric models, e.g., KDEs~\cite{izenman2008modern}. For high dimensional targets, one can use lower dimensional representations. Here, we illustrate our approach on shape sampling and, as in~\cite{tothova2018}, by using a probabilistic PCA prior for $p(y)$
\begin{equation}\label{pca}
y= U S^{\frac{1}{2}}z + \mu + s,
\end{equation}
where $U$ is the matrix of principal vectors, $S$ the diagonal principal component matrix and $\mu$ the data mean. All three were precomputed using surfaces in the training set of $f$. PCA coefficients $z$ and global shift $s$ then modulate and localise the shape in space, respectively. The posterior $p(y|x,\theta,\dset,\model)$ is then approximated by deploying the MH sampling process to estimate the joint posterior of the parameterisation $p(z,s|x,\theta,\dset,\model)$. Proposal shapes $y'$ are constructed at every step of the MCMC from proposals $z'$ and $s'$ for the purposes of likelihood computation as defined by (\ref{eq:MCMC:our_likelihood}) and (\ref{eq:input_loss}). PCA coefficients and shifts are assumed to be independently distributed. We set $p(z) = \mathcal{N}(z;0,I)$ as in~\cite{Bishop2006} and $p(s)=\mathcal{U}([0,D]^2)$ for an input image $x\in\mathbb{R}^{D\times D}$. In practice, we restrict the prior on shift to a smaller area within the image to prevent the accepted shapes from crossing the image boundaries. Assuming both proposals are symmetrical Gaussian, the MH acceptance criterion (\ref{eq:MH}) then becomes $A((z',s'), (z^t,s^t) |x) =  \min \left(1, {\frac{p(x|z',s',f(\cdot;\theta))}{p(x|z^t,s^t,f(\cdot;\theta))}}  \frac{p(z')}{p(z^t)} \frac{p(s')}{p(s^t)}\right)$. 

\section{Results}\label{sec:results}
The method was compared against three uncertainty quantification baselines: (1) \textbf{Post-hoc uncertainty estimation method} RIO~\cite{RIO}, using the code provided by the authors; and  \textbf{Dedicated probabilistic forward systems:} (2) probPCA~\cite{tothova2020} and (3) MC Dropout~\cite{MCDropout}. We compare with MC dropout to provide a complete picture even though it is quantifying epistemic uncertainty.

We evaluate the uncertainty estimates on two regression problems on imaging and non-imaging data. The first is left ventricle myocardium surface delineation in cardiac MR images from the UK BioBank data set \cite{UKBB} where network $f: \mathbb{R}^{D} \times \mathbb{R}^{D} \rightarrow \mathbb{R}^{50 \times 2}$ predicts coordinates of 50 vertices on the surface in small field of view (SFOV) and large FOV (LFOV) images for $D = 60, 200$, respectively. The data sets are imbalanced when it comes to location and orientation of the images with propensity towards the myocardium being in the central area of the image (90\% of the examples). We tested the method on a 10-layer CNN trained to predict a lower dimensional PCA representation from images, as described in Sec.~\ref{sec:prior} and proposed in~\cite{milletari2017}, which is a deterministic version of probPCA.
The CNN predicts 20 PCA components which are then transformed into a surface with 50 vertices. The second is a 1D problem of computed tomography (CT) slice localisation within the body given the input features describing the tissue content via histograms in polar space \cite{dataset_CT_slice,ML_repository_datasets} and $f: \mathbb{R}^{384}\rightarrow [0,100]$. We use the same data set and training setup as in the original RIO paper \cite{RIO} and tested the method on a fully connected network with 2 hidden layers.

The same base architectures were used across baselines, with method pertinent modifications where needed, following closely the original articles~\cite{tothova2018,MCDropout}. No data augmentation was used for forward training. Hyperparameter search for Deep MH was done on the validation set by hand tuning or grid search to get a good balance between the proper optimisation of~(\ref{eq:backprop}), good convergence times and optimal acceptance rates (average of $\sim 20\%$ for high and $\sim 40\%$ for the one-dimensional tasks). This led to $\beta$ = 10000 (SFOV), 30000 (LFOV), 60000 (CT) and Gaussian proposal distributions: $z'\sim\mathcal{N}(z^t, \sigma_z^2 I)$ and $s'\sim\mathcal{N}(s^t, \sigma_s^2 I)$, with ($\sigma_z$, $\sigma_s$) = (0.1, 2) in SFOV, and (0.2, 8) in LFOV; in CT $y'\sim\mathcal{N}(y^t, I)$. Choice of priors for the delineation task followed Sec.~\ref{sec:prior}. In CT, we used an uniformative prior $\mathcal{U}([0,100])$ to reflect that the testing slice can lie anywhere in the body. Parameters of the input backpropagation were set to $\lambda = 1$ and $\sigma_n = 0.1$. We employed independent chain parallelization to speed up sampling and reduce auto-correlation within the sample sets. Chains were initialised randomly and run for a fixed number of steps with initial burn in period removed based on convergence of trace plots of the sampled target variables. The final posteriors were then approximated by aggregation of the samples across the chains. For comparisons, we quantify uncertainty in the system via dispersion of $p(y|x,\theta,\dset,\model)$: standard deviation $\sigma_{y|x}$ in 1D problems; trace of the covariance matrix $\Sigma_{y|x}$ in multi-dimensional tasks (sample covariance matrix for Deep MH and MC Dropout, and predictive covariance for probPCA).

\begin{figure}[t!]
\centering
\begin{minipage}[c]{0.92\columnwidth}
    \includegraphics[angle=0, origin=c, width=\linewidth]{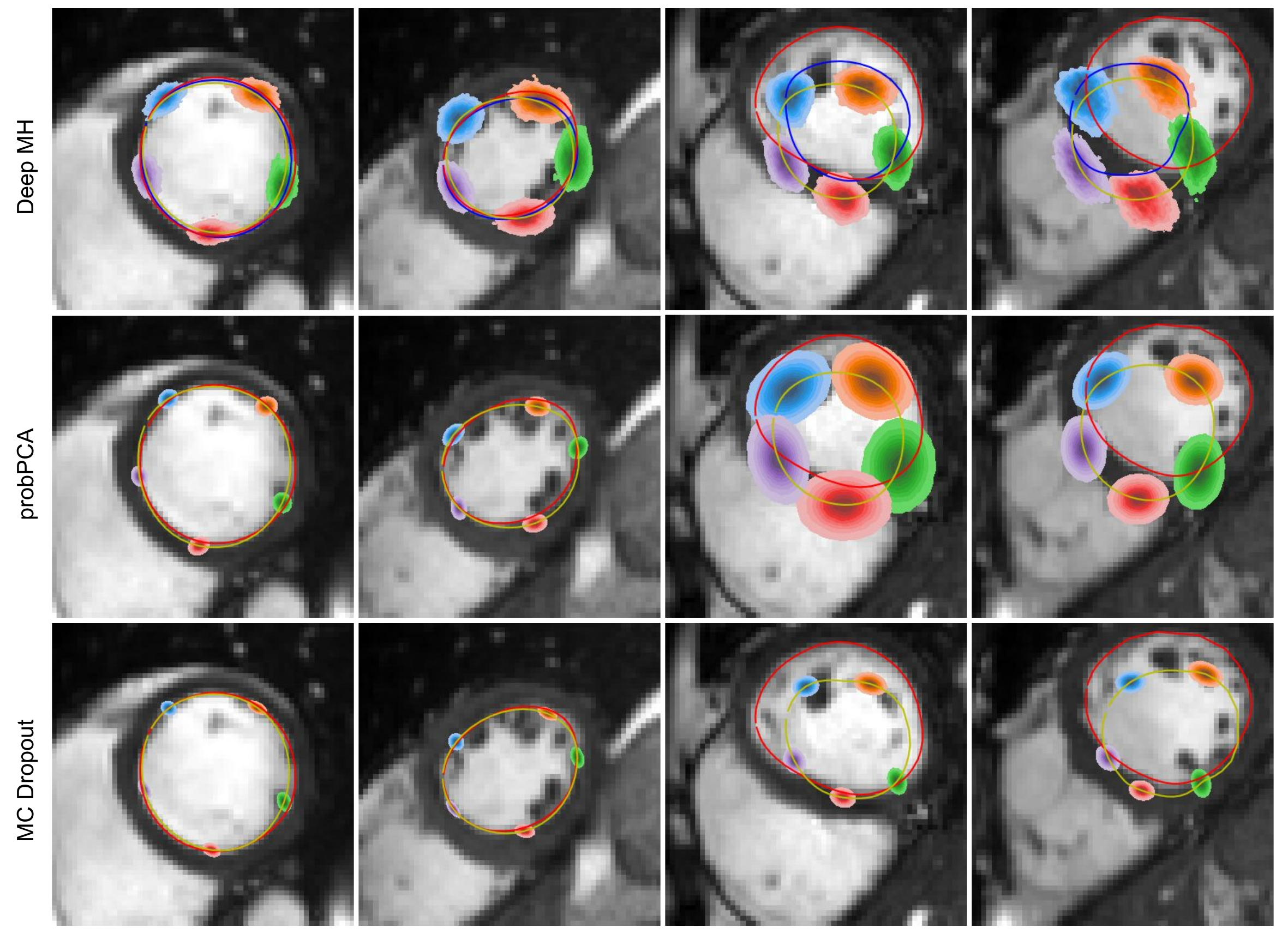}
\end{minipage}
\caption{Comparison of estimated $p(y|x,\theta,\dset,\model)$ for selected test subjects: GT shape (red), forward prediction $f(x_i|\theta)$ (blue), sample / predicted mean (yellow). Kernel density estimates were used for visualization. Only 5 vertices were visualised for clarity. Test subjects are ordered column-wise according to growing fixed network prediction error ($\textrm{RMSE}(f(x),y_{\textrm{GT}}) = 0.48; 2.72; 5.25; 6.31$; test set average $\textrm{RMSE}=1.46$). Outlier test subjects (column 3-4) are associated with higher uncertainty and lower accuracy for all methods.}
\label{fig:samples}
\end{figure}
\noindent\textbf{Predictive Posteriors:}
Fig.~\ref{fig:samples} shows qualitative comparison of target distributions produced by Deep MH, MC Dropout and probPCA in the SFOV surface reconstruction task. We include two images representative of the majority of the data set (well centered) and two from the 10\% outlier group (images with large translation and/or reflection compared to rest of the data set). The posteriors obtained by Deep MH exhibited greater variability, including multimodality, than the baseline methods even for the test images with lower forward prediction RMSE. In contrast, probPCA produced unimodal Gaussians, expectedly. MC Dropout, at $p=0.5$, produced tight distributions around the sample mean.

Deep MH posteriors also showed a larger dispersion of vertex locations along the surface as can be seen in the first two columns. This uncertainty reflects the ambiguity in the target generation process. Boundaries seen in the images can identify the placement of a vertex in the orthogonal direction to the boundary, but placement along the boundary is much less obvious and variable. Deep MH posteriors captured this ambiguity well while others were not able to do so. Irrespective of the method, the high probability regions of the predicted $p(y|x,\theta,\dset,\model)$ might not cover the GT shapes when the prediction fails, as can be seen in the last two columns of Fig.~\ref{fig:samples}.

\noindent\textbf{Correlation with accuracy:} We investigated the relationship between accuracy of a network and the predictive uncertainty. A good uncertainty estimation model should yield a non-decreasing relationship between uncertainty and prediction accuracy, and not show overconfidence, i.e., assign high confidence to imprecise predictions.
\begin{table}[!t]
    \centering
    \begin{tabular}{l|cc|cc}
     & \textbf{Spearman} [r;p] & \textbf{Pearson} [r;p] & \textbf{Spearman} [r;p] & \textbf{Pearson} [r;p]\\
    \hline
    \textbf{Method}& \multicolumn{2}{l|}{\textit{Shape delineation SFOV: full}} & \multicolumn{2}{l}{\textit{Shape delineation SFOV: homogeneous}}\\
    \hline
    Deep MH & $0.30$;  $3 \times 10^{-3}$ & {$0.62$}; {$1\times 10^{-11}$} &{$0.26$}; {$2\times 10^{-2}$} & $0.38$;$3\times 10^{-4}$ \\
    probPCA    & {$0.35$}; {$5 \times 10^{-4}$} &  $0.57$;  $8 \times 10^{-10}$  & $0.19$; $8\times 10^2$  & {$0.40$}; {$9 \times 10^{-5}$}  \\
    MC Dropout &  $0.34$; $5 \times 10^{-4}$    &  $0.61$; $3\times 10^{-11}$& $0.21$;  $5 \times 10^2$  & $0.29$;  $7 \times 10^{-3}$     \\
    \hline
    \textbf{Method}& \multicolumn{2}{l|}{\textit{Shape delineation LFOV: full}} & \multicolumn{2}{l}{\textit{Shape delineation LFOV: homogeneous}}\\
    \hline
    Deep MH       & {$0.33$};  {$3\times 10^{-3}$}       & {$0.38$}; {$7\times 10^{-4}$} & {$0.20$}; {$9 \times 10^{-2}$}   & {$0.26$}; {$3 \times 10^{-2}$} \\
    probPCA    & $0.17$;  $1 \times 10^{-2}$      & $0.33$; $3\times 10^{-3}$& $0.04$; $7 \times 10^{-1}$  & $0.03$; $8 \times 10^{-1}$ \\
    MC Dropout & $0.52$;  $9\times 10^{-7}$ & $0.55$; $2 \times 10^{-7}$  & $0.42$; $2 \times 10^{-4}$  & $0.42$; $2 \times 10^{-4}$  \\
    \hline
    \textbf{Method}&\multicolumn{2}{l|}{\textit{CT slice localisation}}&  \multicolumn{2}{c}{}\\
    \hline
    Deep MH       &{$0.55$};  {$1 \times 10^{-7}$}     &{$0.64$}; {$2\times 10^{-9}$} & \multicolumn{2}{c}{} \\
    RIO        &  $0.07$; $6 \times 10^{-1}$          &  $0.13$; $ 3 \times 10^{-1}$ & \multicolumn{2}{c}{}\\
    MC Dropout &  $0.53$;  $2 \times 10^{-6}$          &  $0.51$; $7 \times 10^{-6}$ &\multicolumn{2}{c}{}\\
    \hline
    \end{tabular}
    \caption{Correlation between uncertainty and accuracy (RMSE).} \label{tab:MCMC:unc_vs_acc}
\end{table}
Tab.~\ref{tab:MCMC:unc_vs_acc} presents the correlation coefficients between quantified uncertainties and the accuracy (RMSE) of either the forward prediction of the pre-trained network in the case of Deep MH, or predictive mean for the baselines. We include results for the shape delineation task analysed in two settings: on the full imbalanced test sets, and on the homogeneous subsets of the test sets with two outliers, detected manually, removed. Deep MH yielded higher Spearman and Pearson's correlation for most cases than probPCA and RIO. Correlation of epistemic uncertainty, as quantified by MC Dropout, with prediction error was higher in some cases than that of aleatoric uncertainty quantified by Deep MH. 

Further analysis together with additional in-depth tests can be found in~\cite{thesis}.

\section{Conclusion}
In this work we proposed a novel method---Deep MH---for uncertainty estimation of trained deterministic networks using MH MCMC sampling. 
The method is architecture agnostic and does not require training of any dedicated NNs or access to training or GT data. 
Experiments on regression tasks showed the better quality of the Deep MH uncertainty estimates not only in comparison to a post-hoc baseline~\cite{RIO}, but dedicated probabilistic forward models as well ~\cite{tothova2018,tothova2020}. 

The main limitation of Deep MH is its computational complexity. Some of it stems from its sampling nature. 
The rest is due to the proposed evaluation of the likelihood. 
A possible alternative to the proposed likelihood computation via optimisation is the simulation of an inversion process via a generative adversarial network (GAN), which would, however, involve a design of an additional dedicated network. 
Deep MH can be applied to problems where it is possible to define or estimate a prior distribution $p(y)$. While the deployment to problems with higher dimensional targets is an open research question, it is straightforward for problems with low effective dimensions. These are commonly found across the spectrum of the deep learning applications. 

\subsubsection{Acknowledgements}
This research has been conducted using the UK Biobank Resource under Application Number 17806.

\bibliography{bibliography}
\end{document}


\title{Contribution Title\thanks{Supported by organization x.}}
%
%
%
\maketitle              
\begin{abstract}
%
%
%


Predictive variability due to data ambiguities has typically been addressed via construction of novel dedicated models with built-in probabilistic capabilities, often trained to predict uncertainty estimates as variables of interest at inference time. These approaches often require distinct architectural components and training mechanisms, include restrictive assumptions and exhibit overconfidence, i.e., high confidence in imprecise predictions.
In this work, we propose a post-hoc sampling strategy for estimating predictive uncertainty accounting for data ambiguity, i.e., aleatoric uncertainty. 
The method can generate different plausible outputs for a given input and does not assume parametric forms of predictive distributions. It is architecture agnostic and can be applied to any feed-forward deterministic network without changes to the network's original architecture nor training procedure. Experiments on regression tasks on imaging and non-imaging input data exemplify the method's ability to generate diverse and multi-model predictive distributions, its generalisability under domain shift and tendency towards avoiding overconfidence compared to alternatives.

\keywords{First keyword  \and Second keyword \and Another keyword.}
\end{abstract}

\section{Introduction}
%
Estimating uncertainty in deep learning (DL) models' predictions, i.e., predictive uncertainty, is of critical importance in a wide range of applications---from medical diagnosis and treatment planning~\cite{Begoli2019} to autonomous driving~\cite{ijcai2017-661} and beyond.
Potential sources of uncertainty in DL prediction systems can come from~\cite{tanno}: (1) \emph{aleatoric} ambiguity in the data or ill-posedness of the task at hand~\cite{WANG1996} that cannot be eliminated by employing larger training sets, or (2) \emph{epistemic} uncertainty~\cite{HORA1996217}) stemming from the model specification, such as parameter uncertainty or model bias~\cite{Draper}, which can be alleviated by training on more data or with a more appropriate model. 

While quantification of predictive uncertainty has been largely addressed by construction of fully-supervised DL models with built-in probabilistic predictions~\cite{ensambles, Neal1996, blundell15, variational_Inference, kendall2017, deep_GP, deep_evidential, tothova2018, MCDropout}, post-hoc uncertainty estimation for pre-trained deterministic networks has been given much less attention.
We believe this avenue is important because most top performing architectures are currently deterministic.
When the DL model is pre-trained and fixed, there is no structural or weight ambiguity, thus, the assessment of predictive uncertainty aims at exploring output variability due to input ambiguity, i.e., aleatoric uncertainty. In the most generic case, this corresponds to characterizing $p(y|x)$, where $x$ and $y$ are input and output, respectively.
This assessment is crucial on its own, setting aside epistemic uncertainty, whenever the nature of the input is inadequate to uniquely identify an output, e.g., measurement noise, bad contrast in computer vision and lack of resolution and partial volume effect in medical image analysis.





Assessment of output variability due to aleatoric uncertainty is challenging. First, the associated posterior distributions can often be complex and intractable.
Second, most real world applications may require generating plausible samples and not just summary statistics~\cite{probunet,phiseg}.  Simplifying approximations of the output variability such as in ~\cite{tothova2020, RIO} might be too limiting. Sampling techniques built on Monte Carlo integration~\cite{monte_carlo} provide a powerful alternative. Traditional Markov Chain Monte Carlo (MCMC) techniques~\cite{Neal93probabilisticinference} construct chains of proposed samples $y'$ with strong theoretical guarantees on their convergence to posterior distribution \pyx{}~\cite{gelman_bayes_book}. In Metropolis-Hastings (MH)~\cite{MH}, this is ensured through evaluation of acceptance probabilities. This involves calculation of a prior $p(y')$---which can be estimated for problems with low dimensional targets, or high dimensional targets with possible lower dimensional representations as we will demonstrate in our experiments---and a likelihood $L(y'; x)= p(x|y')$ at every step.
At its most explicit form, the evaluation of $L(y'; x)$ would translate to the generation of plausible inputs for every given $y'$ and then calculation of the relative density at the input $x$.
If one can sample from or evaluate $p(x,y)$, then this procedure can be followed.
However, in most DL applications, only a finite set of samples from $p(x,y)$ is available. 
The set is used to train a network $f(x;\theta)$ with parameters $\theta$ to predict $y$ given $x$.
In this case where a pre-trained network is available, sampling from $p(y|x,f(\cdot;\theta))$ is still feasible through an appropriate definition of $p(x|y,f(\cdot;\theta))$.
Note that prior work that proposed probabilistic networks define a $p(y|x,f(\cdot;\theta))$ with specific network architectures and parametric distributions, e.g.~\cite{tothova2020}. 
Here we take a different approach and model  $p(x|y,f(\cdot;\theta))$, which allows sampling through an MH scheme.

In some applications, the likelihood function can be defined through analytical energy functionals~\cite{energy_functional1, energy_functional2, energy_functional3}.
In DL, however, we do not have access to a convenient closed form formulation that generates the outputs, which would have allowed an analytical definition.
Alternatively, defining a likelihood with invertible neural networks~\cite{MintNET} or reversible transition kernels specified by NNs~\cite{MC_nets, song} is possible, but these approaches require specialized architectures and training procedures.
For example, the works by Song et al.~\cite{song} and Levy et al.~\cite{MC_nets} use dedicated neural networks in MCMC tasks to design reversible transition kernels. While the former employs adversarial training to devise a likelihood-free transition operator, the latter proposes a family of fast-mixing samplers as a generalization of Hamiltonian Monte Carlo~\cite{Duane1987216, gelman_handbook}.
To the best of our knowledge, a solution that can be applied to any pre-trained network has not been proposed.
Sampling methods that bypass the evaluation of a likelihood do exist, i.e., Approximate Bayesian Computation (ABC)~\cite{ABC:Rubin}, and can be applied to any pre-trained network.
However, they also require to be able to sample from a likelihood, hence the definition of a appropriate likelihood remains a question, see Appendix~\ref{app:ABC} for more details.
We propose a novel MH scheme for sampling of network outputs that can be used for post-hoc evaluation of aleatoric uncertainty.
Inspired by the ideas on network inversion via gradient descent~\cite{image_backprop} and neural style transfer~\cite{style_transfer}, 
the main contribution of our work is a new definition of a likelihood that can be used to evaluate the MH acceptance criterion.
The new likelihood definition uses \textit{input backpropagation} and distances in the input space, motivated by kernel density estimation (KDE)~\cite{izenman2008modern}.
The proposed MH scheme is architecture agnostic, does not require access to training or ground truth data, and avoids explicit formulation of analytically described energy functionals or implementation of dedicated NNs and training procedures.
Thus, it can be readily applied to any pre-trained feed-forward network for which it is possible to define and evaluate a prior $p(y)$. 
It enjoys strong theoretical guarantees on convergence of MCMC, and generates posteriors which can be diverse and multimodal as demonstrated in experiments on real world data sets in Sec.~\ref{sec:results}. Moreover, we observe very desirable uncertainty estimate properties in our experiments on multiple data sets and regression problems.
In comparison to state-of-the-art methods for uncertainty quantification~\cite{MCDropout, tothova2018,RIO}, our method leans towards being cautious when dealing with inaccurate network predictions and tends to reserve high confidence for accurate cases only, which is a very advantageous feature for any predictive purposes.
Domain shift experiments on cardiac magnetic resonance (MR) data show the method's robustness and potential in capturing out of distribution examples as well.

Our experimental evaluation focuses on regression problems, which has been standing a bit outside of the limelight in the uncertainty estimation field. However, the proposed technique can be applied to classification problems without any difficulty.

\section{Related work}
Uncertainty quantification of pre-trained networks has been previously addressed by Qiu et al.~\cite{RIO}. In their work on residual estimation with an I/O kernel (RIO), the authors augment a pre-trained deterministic network by fitting its prediction residuals with a Gaussian process (GP) with a kernel built around the network's inputs and outputs. The GP can then be used for a post-hoc improvement of the network's outputs as well as to assess the predictive uncertainty. This model can be applied to any standard NN without modifications. While mathematically elegant and computationally inexpensive, the straightforward limitations of this work are the requirement for the network outputs to be 1-dimensional, necessary access to the training and ground truth (GT) data, as well as restrictions imposed on the posteriors to be normally distributed. Moreover, as our experiments in Sec.~\ref{sec:results} will show, RIO produces overly confident predictions even when the accuracy is poor.

One of the dedicated DL models addressing aleatoric uncertainty prediction is probPCA~\cite{tothova2018, tothova2020}. Developed for parametric surface reconstruction from images using a principal component analysis (PCA) based shape prior, the method predicts distributions of output shapes as collections of unimodal Gaussian distributions over output mesh vertices. 
This is done using a convolutional NN trained to optimise the distribution of the latent space of the PCA scores using a Kullback–Leibler divergence loss. 

Although not suited for a post-hoc analysis of aleatoric uncertainty, we include Monte Carlo Dropout~\cite{MCDropout} as one of the most popular methods for predictive uncertainty quantification in a wide range of DL tasks. It is a sampling method that uses dropout during training and test time and has been theoretically proven to approximate Bayesian inference in deep GPs.
The first two moments of the sampled set serve as descriptors of the predictive posterior with variance capturing the uncertainty.
The uncertainty estimates are directly tied to the dropout probability $p$ employed and require grid-search to be well calibrated~\cite{concreteDropout}. It has also been shown that MC Dropout might lead to overconfident incorrect predictions on unseen examples~\cite{ensambles}. Despite its widespread adoption, MC Dropout has also been criticised for minimising risk rather than estimating the true uncertainty~\cite{osband}.
\section{Method}\label{sec:MCMC:method}
%
%
We assume a deep neural network $f(x|\theta)$, with $\theta$ representing the parameters, that is pre-trained on a training set of $(x,y)$ pairs, $x \in \mathcal{X} \subset \R^n$, $y \in \mathcal{Y} \subset \R^m$. The network is trained to predict targets from inputs, i.e., $y \approx f(x|\theta)$. We would like to asses the predictive uncertainty of $f(x|\theta)$ induced by aleatoric uncertainty. Ideally, estimating aleatoric uncertainty corresponds to generating samples from $p(y|x)$ for a test input $x$. As this is a challenging task, as detailed in Sec.~\ref{mcmc}, we will use $f(x|\theta)$ to first perform a prediction and then to generate samples with an approximate model. \me{shouldn't all of this be $f(x; \theta)$??}
%
\subsection{Metropolis-Hastings for Target Sampling}\label{mcmc}
The motivation for the proposed method comes from the well established MH MCMC methods for sampling from a posterior distribution~\cite{MH}. 
For a given input $x$ and an initial state $y^0$, the MH algorithm generates Markov chains (MC) of states $\{y^t, t=1, 2, \dots, n\}$. 
At each step $t$ a new proposal is generated $y'\sim g(y'|y^t)$ according to a pre-defined symmetric proposal distribution $g$. 
The sample is then accepted with the probability
\begin{equation}\label{eq:MH}
A(y', y^t |x) =  \min \bigg(1, \underbrace{\frac{g(y^t|y',x)}{g(y'|y^t,x)}}_{\textrm{transitions}} \underbrace{\frac{p(x|y')}{p(x|y^t)}}_{\textrm{likelihoods}}  \underbrace{\frac{p(y')}{p(y^t)}}_{\textrm{priors}}\bigg)
\end{equation}
and the next state is set as $y^{t+1}=y'$ if the proposal is accepted and $y^{t+1}=y^t$ otherwise.
The sufficient condition for asymptotic convergence of the MH MC to a stationary distribution equal to the posterior $p(y|x)$ is satisfied thanks to the reversibility of transitions $y^t\rightarrow y^{t+1}$~\cite{gelman_handbook}.
The asymptotic convergence also means that for arbitrary initialization, the initial samples are not guaranteed to come from $p(y|x)$, hence a burn-in period, where initial chain samples are removed, is often implemented~\cite{gelman_handbook}. Essentially, one can view the MCMC process as gradual addition of chained perturbations to $y^0$.
The goal of the acceptance criterion is then to reject the unfeasible target proposals $y'$ according to prior and likelihood probabilities.
The critical part here is that for every target proposal $y'$, the prior probability $p(y')$ and the likelihood $p(x|y')$ needs to be evaluated. 
While the former is feasible for the problems we focus on
, the latter is not trivial to evaluate.

Evaluation of the likelihood $p(x|y')$ requires construction of a distribution over inputs for a given $y'$.
This is a challenging problem due to the high dimensionality of the inputs and relative small size of data set in the current DL applications.
While research on approximating distributions in high-dimensional spaces with finite data sets is advancing well, and in theory can be applied to estimate $p(x|y')$, e.g.~\cite{kingma2014semi}, we are not aware of any work that used such a model to explore aleatoric uncertainty. More importantly, it is not obvious how one can use such a model to quantify predictive uncertainty of a trained deep neural network.
Here, we take a different approach and characterize a $p(y|x,f(\cdot,\theta))$ that is defined through a novel likelihood model defined in the next section, which makes it applicable to any feed-forward neural network.

\subsection{Likelihood Evaluation}\label{sec:likelihood}
In order to define and characterize a model specific posterior distribution, $p(y|x, f(\cdot;\theta))$, unlike prior work that used a specific network architecture, we define a likelihood function $p(x|y,f(\cdot;\theta)$ that can be used with any pre-trained network.
Given the fixed network $f$ and a proposed target $y'$, 
analogous to~\cite{MC_level_sets, MC_shapes, ertunc, neal2012bayesian}, we define likelihood through the canonical distribution as 
%
\begin{equation}
    \label{eq:MCMC:our_likelihood}
    p(x|y', f(\cdot,\theta)) \propto \exp(-\beta \,E(x,y'))
\end{equation}
where $\beta$ is a "temperature" parameter and $E(x,y')$ is evaluated through a process we call gradient descent \textit{input backpropagation} inspired by neural network inversion approximation in~\cite{image_backprop} and neural style transfer work~\cite{style_transfer}. 

To evaluate the energy for a given $y'$ and $x$, we simulate an input sample $x'_{y'}$ similar to $x$ that could lead to the proposed shape $y' = f(x'_{y'}|\theta)$ without altering the trained network weights. 
This action can be formulated as an optimisation problem 
%
\begin{equation}\label{eq:backprop}
    x'_{y'} = \argminF_{x'} \, \lambda \underbrace{\rho(x', x)}_{\textrm{input loss}} + \underbrace{\mu(f(x'),y')}_{\textrm{target loss}},
\end{equation}
where $\rho: \mathcal{X}\times\mathcal{X} \rightarrow \mathbb{R}^{+}_{0}$, $\mu: \mathcal{Y}\times\mathcal{Y} \rightarrow \mathbb{R}^{+}_{0}$ are distance measures\footnote{Should the problem allow it, a sensible choice for $\mu$ is opting for the original distance measure used during the training of the feed-forward network.} and $\lambda \in \R$ a scaling constant ensuring proper optimisation of both loss elements.
To induce stochasticity, the optimisation is initialised at $x_0\sim\mathcal{N}(x,\sigma_{n}^2 I)$, with $\sigma_n$ denoting the noise level. We then define
\begin{equation}\label{eq:input_loss}
    E(x,y') \coloneqq \rho(x'_{y'}, x).
\end{equation}
We opt for $\rho(\cdot, x)$ equal to squared $L_2$ distance, i.e., $\rho(x'_{y'}, x) = \|x'_{y'} - x\|_2^2$, but other options are possible, as long as (\ref{eq:input_loss}) combined with (\ref{eq:MCMC:our_likelihood}) define a proper probability distribution over $x$. 

It is important to understand Eq.~\ref{eq:backprop} and (\ref{eq:input_loss}) within the MCMC context of the proposed method. The minimisation formulation ensures that we only generate $x'_{y'}$ close to test image $x$ that are the most likely to lead to the proposed $y'$.
Provided the two terms in (\ref{eq:backprop}) are balanced correctly, a low likelihood value assigned to an MCMC proposal $y'$ indicates the output either lies outside the range of $f$ (the original fixed feed-forward network is incapable of producing this kind of shape for any kind of input), or outside of the image (understood as a subset of codomain of $f$) of the inputs similar to $x$ under $f$\footnote{Just as in ABC, the exact definition of similarity depends on the pre-selected distance measure $\rho$.}.

The choice of $\beta$ parameter in (\ref{eq:MCMC:our_likelihood}) is of particular importance.
It affects the acceptance rate and plays a crucial role during the evaluation of the acceptance criterion in MH sampling where it prevents catastrophic numerical cancellation in cases where $E(x,y')$ and $p(y')$ have disproportionately different orders of magnitude in the negative log scale used in the computations\footnote{Logarithmic scale is a typical choice used to avoid issues such as overflow.}. As a result, the acceptance ratio is not dominated by either one. Fig.~\ref{fig:backprop_scheme} illustrates the optimisation setup.
The optimisation problem (\ref{eq:backprop}) needs to be solved for every MCMC proposal $y'$. The full proposed MH MCMC sampling scheme is described in Algorithm~\ref{MCMC:alg:whole_method}. We will refer to it as \textbf{Deep MH}. Thanks to the formulation of the likelihoods (\ref{eq:MCMC:our_likelihood}) as compound distributions (see Appendix~\ref{app:compound_dist}), the resulting sampling process allows for the posterior distributions $p(y|x,f(\cdot,\theta))$ to be highly complex and multimodal, as confirmed by the experiments in Sec.~\ref{sec:results}.
%

\paragraph{Links to \bm{$p(x|y')$}} The motivation for using the distribution given in Eq.~\ref{eq:MCMC:our_likelihood} and (\ref{eq:input_loss}) comes from \textit{kernel density estimation (KDE)}~\cite{izenman2008modern}. A kernel density estimation of $p(x|y')$ can be written as 
\begin{equation*}
    p(x|y') \approx \frac{1}{N}\sum_{k=1}^N K_h(x - x'_k),\quad x'_k\sim p(x|y')
\end{equation*}
where $K_h(\cdot)$ is the kernel with $h$ its bandwidth. If the kernel function is always positive and integrates to 1 then the KDE approximation is asymptotically consistent without any restrictions on $p(x|y')$ and unbiased under some regularity conditions~\cite{izenman2008modern}.

The KDE requires samples from the posterior distribution $p(x|y')$. In practical problems, we do not necessarily have $N$ samples from $p(x|y')$ for arbitrary $y'$.
Considering finite data sets, we may not even have any samples from these distributions.
Instead, we assume that a trained and fixed network can be used to generate $x$ samples for a given $y'$, and one way to do this is to solve the optimisation problem given in Eq.~\ref{eq:backprop}.
The choice of the energy function is also not random. Eq.~\ref{eq:input_loss} with the $L_2$ distance and $\beta$ corresponds to a Gaussian kernel with bandwidth equal to $\sqrt{1/\beta}$.

Lastly, one can generate many $x'_{y'}$ by randomly initialising the optimisation.
These samples can then be summed for the KDE approximation, i.e., $p(x|y') \approx 1/N\sum K(x-x'_{y'}{}^{(k)})$ where $x'_{y'}{}^{(k)}$ are solutions to the optimisation problem with different initialisation.
In case only one sample is generated, we use only one kernel evaluation instead of the sum, which reduces the computational load, and gives rise to the definition in Eq.~\ref{eq:MCMC:our_likelihood}.
We acknowledge that this distribution is an approximation to $p(x|y')$, but one of the only practical avenues to the best of our knowledge.
%
\begin{figure}
    \centering
    \includegraphics[width=0.9\columnwidth]{chapters/MCMC/figures/input_backprop.pdf}
    \caption{Depiction of the input backpropagation process. Given a pre-trained network $f$, a test image $x$ and a target proposed shape $y'$, we optimise for an image $x'$ via backpropagation. Note that we initialise $x'$ at $x$ corrupted with Gaussian noise. The backpropagation process is repeated until all the components of the loss~(\ref{eq:backprop}) are minimised.}
    \label{fig:backprop_scheme}
\vskip -0.15in
\end{figure}
%

\begin{algorithm}[t]
\caption{Deep MH target sampling using input backpropagation}\label{algorithm}\label{MCMC:alg:whole_method}
\begin{algorithmic}
\begin{small}
\STATE{\textbf{input:} test image $x$, proposal distribution $g$, pre-trained network $f$, initial shape $y^0$}
\STATE {$t \leftarrow 0$;}
\REPEAT \STATE{
    $y' \sim g(y'|y^t)$;\\
    $ E(x, y')  \leftarrow$ input loss \me{$\rho(x'_{y'}, x)$} (\ref{eq:backprop}) (see also Fig.~\ref{fig:backprop_scheme});\\
    $p(x|y') \leftarrow $ likelihood (\ref{eq:MCMC:our_likelihood});\\
    $A(y', y^t|x) \leftarrow $ acceptance probability (\ref{eq:MH});\\
    $u\sim\mathcal{U}([0,1])$;\\
    \IF{$u \leq A(y' , y^t|x)$}\STATE{
        accept and set $y^{t+1} \leftarrow y'$;}
    \ELSE \STATE{
        reject and set $y^{t+1} \leftarrow y^t$;\\
    }
    \ENDIF \\
    \STATE {$t \leftarrow t + 1$;}
}
\UNTIL {desired chain length}
\end{small}
\end{algorithmic}
\end{algorithm}

\subsection{MH for Shape Sampling Using Lower Dimensional Parametric Representations}\label{sec:prior}
In certain applications direct sampling of the 
targets $y$ might not be feasible due to the high dimensionality of the target space or the need to embed a more complex prior knowledge into the sampling procedure.
As in~\cite{tothova2018} we look at the specific case of a lower dimensional parametric shape representation $y(z,s)$ using a pre-trained PCA prior
\begin{equation}\label{pca}
y= U S^{\frac{1}{2}}z + \mu + s,
\end{equation}
where $U$ is a column-wise matrix of principal vectors, $S$ the principal component diagonal covariance matrix and $\mu$ the data mean. All three were precomputed using surfaces in the training set of $f$. PCA coefficients $z$ and global shift $s$ then modulate and localise the shape in space, respectively. The posterior $p(y|x,f(\cdot;\theta))$ is then approximated by deploying the MH sampling process to estimate the joint posterior of the parameterisation $p(z,s|x,f(\cdot;\theta))$. Proposal shapes $y'$ are constructed at every step of the MCMC from proposals $z'$ and $s'$ for the purposes of likelihood computation as defined by (\ref{eq:MCMC:our_likelihood}) and (\ref{eq:input_loss}). PCA coefficients and shifts are assumed to be independently distributed. The prior on $z\sim\mathcal{N}(0,I)$ stems from probabilistic PCA~\cite{Bishop2006} and restricts the reconstructed shapes to lie on a manifold of feasible shapes given the particular training set. As the shape can lie anywhere in the image, we set the prior $s\sim\mathcal{U}([0,D]^2)$ for an input image $x\in\mathbb{R}^{D\times D}$. In practice, the prior on shift is restricted to a smaller area within the image to prevent the accepted shapes from crossing the image boundaries. Assuming both proposals are symmetrical Gaussian, the MH acceptance criterion (\ref{eq:MH}) then becomes $A((z',s'), (z^t,s^t) |x) =  \min \left(1, {\frac{p(x|z',s',f(\cdot;\theta))}{p(x|z^t,s^t,f(\cdot;\theta))}}  \frac{p(z')}{p(z^t)} \frac{p(s')}{p(s^t)}\right)$. 

\section{Results}\label{sec:results}

The method was compared against three uncertainty quantification baselines: (1) \textbf{Post-hoc uncertainty estimation method} RIO~\cite{RIO}, using the code provided by the authors; and  \textbf{Dedicated probabilistic forward systems:} (2) probPCA~\cite{tothova2020} and (3) MC Dropout~\cite{MCDropout}. Note that we chose to compare with MC dropout to provide a complete picture even though it is quantifying epistemic uncertainty.



We evaluate the uncertainty estimates on two types of real-world regression problems on imaging and non-imaging data of varying feature dimensions and data set sizes: 
(1) Left ventricle myocardium surface delineation in cardiac MR images from the UK BioBank data set \cite{UKBB} where network $f: \R^{D} \times \R^{D} \rightarrow \R^{50 \times 2}$ predicts coordinates of 50 vertices on the surface in small field of view (SFOV) and large FOV (LFOV) images for $D = 60, 200$, respectively. The data sets are imbalanced when it comes to location and orientation of the images with propensity towards the myocardium being in the central area of the image (roughly 90\% of the examples). The method was tested on a trained 10-layer CNN using a lower dimensional PCA representation, as described in Sec.~\ref{sec:prior} and proposed in~\cite{milletari2017}, which is a deterministic version of probPCA.
The CNN predicts 20 PCA components which are then transformed into a surface with 50 vertices.\\
(2) A one-dimensional problem of computed tomography (CT) slice localisation within the body given the input features describing the tissue content via histograms in polar space \cite{dataset_CT_slice, ML_repository_datasets} and $f: \R^{384}\rightarrow [0,100]$. We use the same data set and training setup as in the original RIO paper \cite{RIO} and tested the method on a fully connected network with 2 hidden layers.

The same base architectures were used across baselines, with method pertinent modifications where needed, following closely the original articles~\cite{tothova2018, MCDropout}. No data augmentation was used for forward training. Hyperparameter search was done on the validation set by hand tuning or grid search. The focus of calibration of the proposed Deep MH  were proper optimisation of both loss elements during input backpropagation~(\ref{eq:backprop}), good convergence times and optimal acceptance rates (average of $\sim 20\%$ for high and $\sim 40\%$ for the one-dimensional tasks). Combined with restrictions on $\beta$ from Sec. \ref{sec:likelihood}, this leads to a very specific choice of the method's parameters $\beta$ = 10000 (SFOV), 30000 (LFOV), 60000 (CT) and proposal step sizes. We used Gaussian proposal distributions: for shape delineation $z'\sim\mathcal{N}(z^t, \sigma_z^2 I)$, $s'\sim\mathcal{N}(s^t, \sigma_s^2 I)$, with ($\sigma_z$, $\sigma_s$) = (0.1, 2) in SFOV, and (0.2, 8) in LFOV; in CT $y'\sim\mathcal{N}(y^t, I)$. Choice of priors for the delineation task followed Sec.~\ref{sec:prior}. In CT, we used an uniformative prior $\mathcal{U}([0,100])$ to reflect that the testing slice can lie anywhere in the body. Parameters of the input backpropagation were set to $\lambda = 1$ and $\sigma_n = 0.1$. See Appendix~\ref{app:experiment_setup} for details on experimental and training setup, data sets, and specific parameter values used. Further analysis of the role of the acceptance rate can be found in Appendix~\ref{sec:MCMC:acc_rate}. We employed independent chain parallelisation to speed up sampling and reduce auto-correlation within the sample sets. Chains were initialised at randomly dispersed locations and run for a fixed number of steps with initial burn in period removed
%
based on convergence of trace plots of the sampled target variables. The final posteriors were then approximated by aggregation of the samples across the independent converged chains.

For comparisons, we quantify uncertainty in the system via dispersion of $p(y|x,f(\cdot;\theta))$: standard deviation $\sigma_{y|x}$ in 1D problems; trace of the covariance matrix $\Sigma_{y|x}$ of $p(y|x,f(\cdot;\theta))$ in multi-dimensional tasks (sample covariance matrix for Deep MH and MC Dropout, and predictive covariance for probPCA).
%
 \begin{figure}[t!]
\centering
\begin{minipage}[c]{0.95\columnwidth}
    \includegraphics[angle=0, origin=c, width=\linewidth]{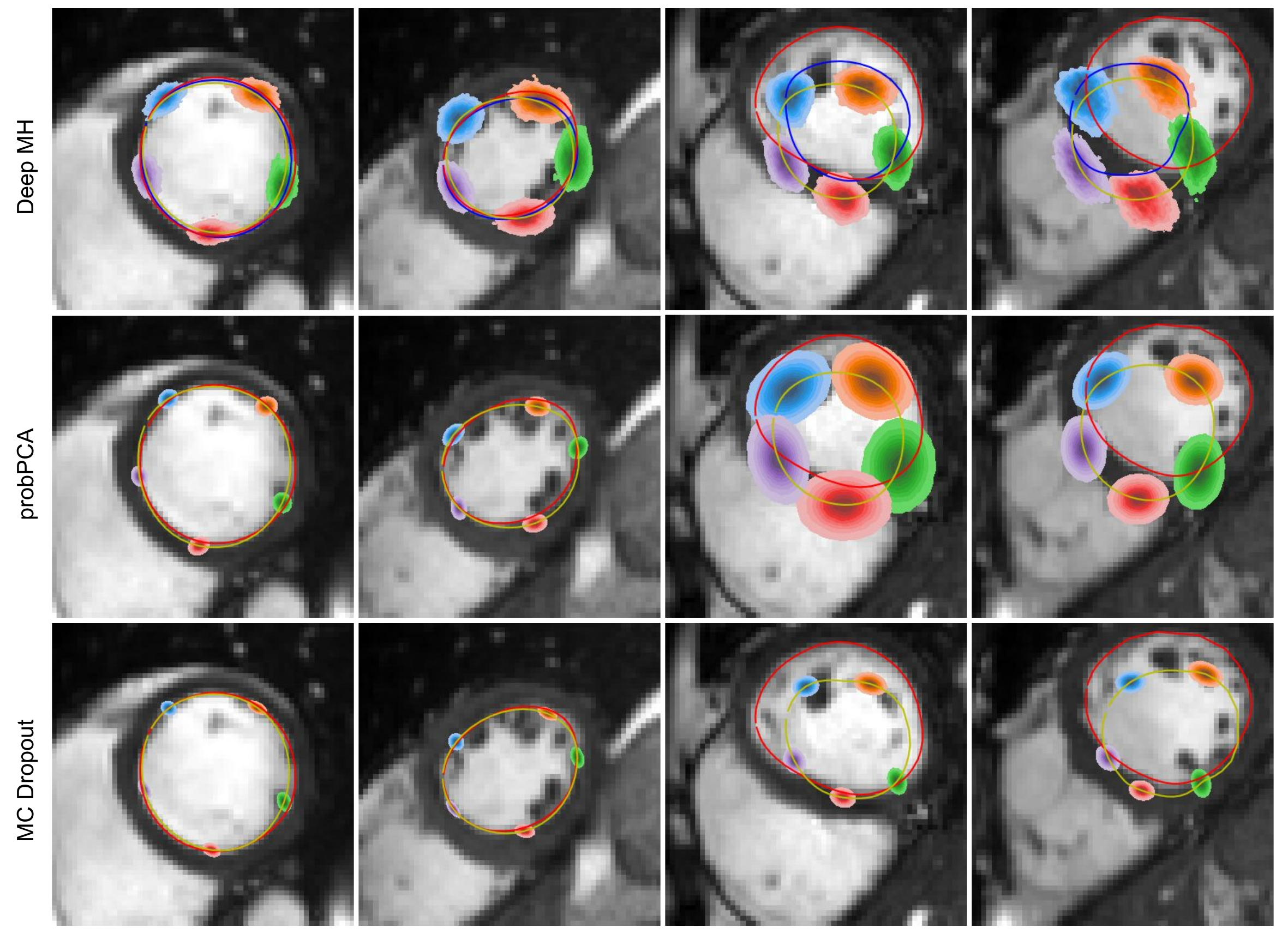}
\end{minipage}
\caption{Comparison of estimated $p(y|x,f(\cdot;\theta))$ for selected test subjects: GT shape (red), forward prediction $f(x_i)$ (blue), sample / predicted mean (yellow). Kernel density estimates were computed using 90000 samples using a Gaussian kernel. Only 5 vertices were visualised for clarity. Test subjects are ordered column-wise according to growing fixed network prediction error ($\textrm{RMSE}(f(x),y_{\textrm{GT}}) = 0.48; 2.72; 5.25; 6.31$; test set average $\textrm{RMSE}=1.46$). Outlier test subjects (column 3-4) are associated with higher prediction uncertainty and accuracy across the methods. Predictive error of the baseline methods is not included, as accuracy improvement is out of the scope of this work.
}
\label{fig:samples}
\end{figure}
%
\paragraph{Predictive Posteriors} 
Fig.~\ref{fig:samples} shows qualitative comparison of target distributions produced by Deep MH, MC Dropout and probPCA in the SFOV surface reconstruction task.
We include two images representative of the majority of the data set (well centered) and two from the 10\% outlier group (off center). The posteriors obtained by our scheme exhibit greater variability, including multimodality, than the baseline methods even for the test images with lower forward prediction RMSE. In contrast, even at the dropout probability $p=0.5$, MC Dropout produces very tight distributions around the sample mean
while the probPCA shows variability in terms of variance, the predictive distribution is a unimodal Gaussian, as per the assumptions of the method. Samples corresponding to the visualised distributions are shown in the Appendix~\ref{app:results}.

We also observe that deep MH posteriors show a larger dispersion of vertex locations along the surface as can be seen in the first two columns. This uncertainty reflects the ambiguity in the target generation process well. Boundaries seen in the images can identify the placement of a vertex in the orthogonal direction to the boundary, but placement along the boundary is much less obvious and variable. Vertex locations along the boundary are established with approximate correspondence between shapes as explained 
in Appendix~\ref{sec:datasets}. Deep MH posteriors capture this ambiguity well while probPCA and MC Dropout were not able to do so.

Irrespective of the method, the high probability regions of the predicted $p(y|x,f(\cdot;\theta))$ might not cover the GT shapes when the prediction fails, as can be seen in the last two columns of Fig.~\ref{fig:samples}.
For deep MH, this may be due to (i) the limited ability of the fixed network to ever achieve the expert-defined GT shape, either because it cannot be recovered from the limited number of PCA components or because the shape was not well represented in the training set, (ii) the insufficient capacity of the network $f$, or (iii) insufficient exploration of the posterior. 

To investigate the possibility that this is happening due to insufficient exploration of the posterior space
in Deep MH, for a case with high RMSE, we initialised 2 chains randomly in the neighbourhood of the 
$(z,s)$ for the forward prediction $f(x)$ and 3 chains around the $(z,s)$ values for the pertinent GT shape $y_{\textrm{GT}}$. This was done for analysis purposes only.
Deep MH, and in general MCMC sampling, does not require access to the ground truth for test samples.
Fig.~\ref{fig:mixing} shows sampled locations for 5 vertices of the recorded Deep MH chains.
Even though the two groups of chains start off at distinctly different values, they converge to a common point after the burn in.
This shows that the issue is not with insufficient exploration of the posterior but the GT shapes indeed do not lie in high probability regions of the posterior. Further convergence properties of Deep MH are explored in Appendix~\ref{app:convergence}.
\begin{figure}[t!]
\centering
\begin{minipage}[c]{1\columnwidth}
	\includegraphics[ angle=0, origin=c, width=\linewidth]{chapters/MCMC/figures/MCMC/mixing/new/trace_vertex_1024925_cropped_2.pdf}
\end{minipage}
\\
\begin{minipage}[c]{1\columnwidth}
	\includegraphics[angle=0, origin=c, width=\linewidth]{chapters/MCMC/figures/MCMC/mixing/new/vertex_trace_dim2_cropped.pdf}
\end{minipage}
\caption{Convergence behaviour of the Deep MH process with 5 chains for subject with poor forward prediction $f(x)$ via trace plots of coordinates of selected vertices (one dimension per row), all corresponding to 3rd subject in Fig.~\ref{fig:samples}. Three chains were initialised in the neighbourhood of the GT reference shape, 2 around $f(x)$. Irrespective of the initialisation, the chains converge to the same equilibrium. Traces of corresponding $s$ and selected elements of $z$ are pictured in Fig.~\ref{fig:app:mixing} in the Appendix.}\label{fig:mixing}
\vskip -0.20in
\end{figure}

\paragraph{Accuracy as a Function of Uncertainty} One of the driving forces behind the development of uncertainty estimation methods is their potential to assist in DL-based decision making. 
A successful uncertainty method is able to associate inaccurate predictions with high uncertainty, and defer decisions in such cases to a specialist.
With this motivation we investigate the relationship between accuracy of the forward network and the predictive uncertainty.
What we would like to see is a non-decreasing relationship between uncertainty and prediction accuracy.
What we would ideally not want to observe is overconfidence, i.e. assignment of relatively high confidence to imprecise predictions.
\begin{figure}[t!]
\centering
\begin{minipage}{\columnwidth}
\begin{minipage}[t]{0.3\columnwidth}
	\includegraphics[angle=0, origin=r, width = \columnwidth]{chapters/MCMC/figures/CT_localisation/uncertainty/RIO_uncertainty_vs_rmse.pdf}
\end{minipage}
\hfill
\begin{minipage}[t]{0.3\columnwidth}
	\includegraphics[angle=0, origin=r, width = \columnwidth]{chapters/MCMC/figures/CT_localisation/uncertainty/MC Dropout_uncertainty_vs_rmse.pdf}
\end{minipage}
\hfill
\begin{minipage}[t]{0.3\columnwidth}
	\includegraphics[angle=0, origin=r, width = \columnwidth]{chapters/MCMC/figures/CT_localisation/uncertainty/Ours_uncertainty_vs_rmse.pdf}
\end{minipage}
\end{minipage}
\\
\begin{minipage}{\columnwidth}
\begin{minipage}[t]{0.3\columnwidth}
	\includegraphics[angle=0, origin=r, width = \columnwidth]{chapters/MCMC/figures/uncertainty/small_plots/probPCA_uncertainty_vs_rmse.pdf}
\end{minipage}
\hfill
\begin{minipage}[t]{0.3\columnwidth}
	\includegraphics[angle=0, origin=r, width = \columnwidth]{chapters/MCMC/figures/uncertainty/small_plots/MC Dropout_uncertainty_vs_rmse.pdf}
\end{minipage}
\hfill
\begin{minipage}[t]{0.3\columnwidth}
	\includegraphics[angle=0, origin=r, width = \columnwidth]{chapters/MCMC/figures/uncertainty/small_plots/ours_uncertainty_vs_rmse.pdf}
\end{minipage}
\end{minipage}
\\
\begin{minipage}{\columnwidth}
\begin{minipage}[t]{0.3\columnwidth}
	\includegraphics[angle=0, origin=r, width = \columnwidth]{chapters/MCMC/figures/uncertainty/small_plots/largeFOV/probPCA_uncertainty_vs_rmse.pdf}
\end{minipage}
\hfill
\begin{minipage}[t]{0.3\columnwidth}
	\includegraphics[angle=0, origin=r, width = \columnwidth]{chapters/MCMC/figures/uncertainty/small_plots/largeFOV/MC Dropout_uncertainty_vs_rmse.pdf}
\end{minipage}
\hfill
\begin{minipage}[t]{0.3\columnwidth}
	\includegraphics[angle=0, origin=r, width = \columnwidth]{chapters/MCMC/figures/uncertainty/small_plots/largeFOV/ours_uncertainty_vs_rmse.pdf}
\end{minipage}
\end{minipage}
\caption{Accuracy as function of uncertainty over test set. Top to bottom: CT slice localisation, Shape delineation full SFOV, Shape delineation full LFOV.}
\label{fig:MCMC:res:unc_vs_acc}
\vskip -0.2in
\end{figure}

\begin{table}[t!]
\begin{small}
\caption{Correlation between uncertainty and accuracy (RMSE).} \label{tab:MCMC:unc_vs_acc}
\begin{center}
\begin{tabular}{lcc}
\hline
\textbf{Method} & \textbf{Spearman} [r;p] & \textbf{Pearson} [r;p]  \\
\hline
    \multicolumn{3}{l}{\textit{Shape delineation SFOV: full}}  \\
\hline
    Ours       &  $0.30$;  $3 \times 10^{-3}$      & \bm{$0.62$}; \bm{$1\times 10^{-11}$} \\
    MC Dropout &  $0.34$; $5 \times 10^{-4}$    &  $0.61$; $3\times 10^{-11}$     \\
    probPCA    & \bm{$0.35$}; \bm{$5 \times 10^{-4}$} &  $0.57$;  $8 \times 10^{-10}$     \\
\hline
\multicolumn{3}{l}{\textit{Shape delineation SFOV: homogeneous}} \\
\hline
    Ours      &\bm{$0.26$}; \bm{$2\times 10^{-2}$} & $0.38$;$3\times 10^{-4}$ \\
    MC Dropout & $0.21$;  $5 \times 10^2$  & $0.29$;  $7 \times 10^{-3}$ \\
    probPCA    & $0.19$; $8\times 10^2$  & \bm{$0.40$}; \bm{$9 \times 10^{-5}$}\\
\hline
\multicolumn{3}{l}{\textit{Shape delineation LFOV: full}}   \\
\hline
    Ours       & $0.33$;  $3\times 10^{-3}$       & $0.38$; $7\times 10^{-4}$  \\
    MC Dropout & \bm{$0.52$};  \bm{$9\times 10^{-7}$} & \bm{$0.55$}; \bm{$2 \times 10^{-7}$}    \\
    probPCA    & $0.17$;  $1 \times 10^{-2}$      & $0.33$; $3\times 10^{-3}$ \\
\hline
\multicolumn{3}{l}{\textit{Shape delineation LFOV: homogeneous}} \\
\hline
    Ours      & $0.20$; $9 \times 10^{-2}$   & $0.26$; $3 \times 10^{-2}$\\
    MC Dropout & \bm{$0.42$}; \bm{$2 \times 10^{-4}$}  & \bm{$0.42$}; \bm{$2 \times 10^{-4}$} \\
    probPCA    & $0.04$; $7 \times 10^{-1}$  & $0.03$; $8 \times 10^{-1}$ \\
\hline
\multicolumn{3}{l}{\textit{CT slice localisation}} \\
\hline
    Ours       &\bm{$0.55$};  \bm{$1 \times 10^{-7}$}     &\bm{$0.64$}; \bm{$2\times 10^{-9}$} \\
    MC Dropout &  $0.53$;  $2 \times 10^{-6}$          &  $0.51$; $7 \times 10^{-6}$ \\
    RIO        &  $0.07$; $6 \times 10^{-1}$          &  $0.13$; $ 3 \times 10^{-1}$  \\
\hline
\end{tabular}
\end{center}
\end{small}
\vskip -0.3in
\end{table} 
Fig.~\ref{fig:MCMC:res:unc_vs_acc} relates the uncertainties obtained for each test image to the accuracy (RMSE) of either the forward prediction of the pre-trained network in the case of Deep MH, or predictive mean for the baselines. The pertinent correlation coefficients can be found in Tab.~\ref{tab:MCMC:unc_vs_acc}, which also includes the results for shape delineation task analysed in two settings: on the full imbalanced test sets, and on the homogeneous subsets of the test sets with the outlier values removed. These were identified manually based on the orientation and displacement of the organ within the image. Across the data sets, the baselines displayed a propensity towards assigning high uncertainty to predictions with low accuracy.
Our method leaned towards being more cautious across the data set with lower covariance trace almost exclusively reserved for samples with lower RMSE and delivered better correlation on the CT and SFOV data sets.
The post-hoc estimation method RIO failed to exhibit desirable behaviour.
\begin{figure*}[t!]
\centering
\begin{minipage}[t]{0.4\columnwidth}
	\includegraphics[angle=0, origin=c, width=\columnwidth]
	{chapters/MCMC/figures/uncertainty/conf_plots/CT/acc_vs_confidence_max_.pdf}
\subcaption{CT}
\end{minipage}
\\
\begin{minipage}[t]{0.4\columnwidth}
	\includegraphics[angle=0, origin=c, width=\columnwidth]
	{chapters/MCMC/figures/uncertainty/conf_plots/smallFOV/acc_vs_confidence_max_.pdf}
	\subcaption{Full SFOV}
\end{minipage}
\hfill
\begin{minipage}[t]{0.4\columnwidth}
	\includegraphics[angle=0, origin=c, width=\columnwidth]
	{chapters/MCMC/figures/uncertainty/conf_plots/smallFOV/acc_vs_confidence_max_ODexRemoved.pdf}
\subcaption{Hom. SFOV}
\end{minipage}
\\
\begin{minipage}[t]{0.4\columnwidth}
	\includegraphics[angle=0, origin=c, width=\columnwidth]
	{chapters/MCMC/figures/uncertainty/conf_plots/largeFOV/acc_vs_confidence_max_.pdf}
	\subcaption{Full LFOV}
\end{minipage}
\hfill
\begin{minipage}[t]{0.4\columnwidth}
	\includegraphics[angle=0, origin=c, width=\columnwidth]
	{chapters/MCMC/figures/uncertainty/conf_plots/largeFOV/acc_vs_confidence_max_ODexRemoved.pdf}
	\subcaption{Hom. LFOV}
\end{minipage}
\caption{Accuracy as a function of uncertainty. Confidence plots of scaled relative uncertainty. All the baselines show the propensity to producing incorrect yet confident predictions.}
\label{fig:MCMC:res:acc_confplots_}
\end{figure*}

Correlation coefficients
cannot capture a method's "readiness for overconfidence". 
To compare the methods' inclination to attribute high confidence to incorrect predictions, we propose a novel regression uncertainty evaluation tool inspired by the confidence plots developed by Lakshminarayanan et al.~\cite{ensambles} for classification problems.
Here, the model's accuracy is evaluated on the cases where its confidence is above a given threshold. The calculated accuracy is then plotted against the thresholded confidence. 
In regression, the prediction range, the accuracy, and---as observed in Fig.~\ref{fig:MCMC:res:unc_vs_acc}---uncertainty are unconstrained and variable between the methods\footnote{The discrepancy in the absolute scale of the predictive uncertainty obtained by different methods is itself a point for further exploration beyond the scope of this work.}. 
To ensure a fair comparison of the patterns of predictive uncertainty, we rescaled the accuracy (measured by RMSE) and uncertainty of the test examples onto [0,1]. Examples whose scaled uncertainty is below a given threshold $0<\tau\leq1$ are then filtered and their minimum accuracy (i.e. max RMSE over the subset at this uncertainty threshold) plotted. We expect the plots to be monotonically non-decreasing with the curve of the least over-confident method lying below the others, especially at the lower end of $\tau$ range corresponding to high confidence situations.
The most prudent method can ensure high accuracy of the most confident predictions, and it will be comparatively slow in increasing its confidence as the predictive error increases.
In other words, max RMSE would increase slowly with increasing $\tau$.
The results can be seen in Fig.~\ref{fig:MCMC:res:acc_confplots_}.
Deep MH exhibits the most balanced and desirable relationship between uncertainty and accuracy across the data sets.
RIO's bad performance in this test was due to assigning low confidence in general, see rescaled scatter plots in Fig.~\ref{fig:MCMC:res:acc_confplots} in the Appendix~\ref{app:results}.
The experiment also confirmed the results reported in~\cite{ensambles}: MC Dropout can produce overconfident wrong predictions. ProbPCA fares similarly.
\paragraph{Generalisation under Domain Shift} The analysis so far involved test sets $p_{\textrm{TEST}}(x,y)$ sampled from the same distribution as the training set $p_{\textrm{TRAIN}}(x,y)$. 
The domain shift setting evaluates the uncertainty quality on out-of-distribution examples from modified input distributions $p^{*}_{\textrm{TEST}}(x,y) \neq p_{\textrm{TRAIN}}(x,y)$~\cite{dataset_shift}. 
We induce an input covariate shift in $p(x)$ in the SFOV task by progressive corruption of the original test set by noise (i.e. input data augmented with Gaussian noise  $\sim\mathcal{N}(0,\sigma^{2} I)$, for an increasing $\sigma$) and rotation by an increasing angle $\alpha$. 
The desirable behaviour of high quality uncertainty estimates is that the further the sample from the original distribution, the higher the attributed uncertainty should be.
The results in Fig.~\ref{fig:MCMC:OOD} show our method conforms with this expectation.
%
\begin{figure}[]
\centering
\begin{minipage}[c]{0.99\columnwidth}
\centering
    \begin{minipage}[c]{0.42\columnwidth}
    	\includegraphics[angle=0, origin=c, width=\columnwidth]{chapters/MCMC/figures/OOD/Noise/scaled/noise_vs_tr(cov)_normRMSE_normUNC.pdf}
    \end{minipage}
    \begin{minipage}[c]{0.42\columnwidth}
    	\includegraphics[ angle=0, origin=c, width=\columnwidth]{chapters/MCMC/figures/OOD/Noise/scaled/noise_vs_RMSE_normRMSE_normUNC.pdf}
    \end{minipage}
\end{minipage}
\\
\begin{minipage}[c]{0.99\columnwidth}
\centering
    \begin{minipage}[c]{0.42\columnwidth}
    	\includegraphics[angle=0, origin=c, width=\columnwidth]{chapters/MCMC/figures/OOD/Rotation/scaled/rotation_vs_tr(cov)_normRMSE_normUNC.pdf}
    \end{minipage}
    \begin{minipage}[c]{0.42\columnwidth}
    	\includegraphics[ angle=0, origin=c, width=\columnwidth]{chapters/MCMC/figures/OOD/Rotation/scaled/rotation_vs_RMSE_normRMSE_normUNC.pdf}
    \end{minipage}
\end{minipage}
\caption{Uncertainty under domain shift due to noise (top) and rotation (bottom). Values scaled for comparison between methods onto [0,1]. For computational reasons, the experiments were carried out on the task of surface using a test subset (3 random subjects). The predicted uncertainty increases then drops slightly after $-120^{\circ}$. This might be due to the models picking up on the symmetry relationships between $p^{*}_{\textrm{TEST}}(x,y)$ and the original unrotated $p_{\textrm{TRAIN}}(x,y)$ at each rotation level, with $-120^{\circ}$ being the most dissimilar.}\label{fig:MCMC:OOD}
\vskip -0.28in
\end{figure}
We observe an almost linear relationship between the predicted uncertainty and the noise level for Deep MH.
While the forward networks exhibit robustness to disruptions in the data up to a certain point, 
Deep MH picks up on the changes in the input distribution by increasing the uncertainty at the lowest level compared to baselines.
For rotations, the predictive uncertainty across methods exhibits a general increasing trend for increasing $\alpha$; the uncertainty performance of Deep MH is comparable to the baselines.
Visualisations of the $p^{*}_{\textrm{TEST}}(x,y)$ and predictive posteriors can be found in Appendix~\ref{app:results} Fig.~\ref{fig:app:OOD}.

\section{Conclusion}
In this work we proposed a novel method---Deep MH---for uncertainty estimation of trained deterministic networks using MH MCMC sampling. 
We used an approximation to the likelihood evaluation motivated by kernel density estimation. 
The method is architecture agnostic and does not require training of any dedicated NNs or access to training or GT data. 
Experiments on small and high dimensional imaging and non imaging regression tasks show the better quality of the Deep MH uncertainty estimates not only in comparison to a post-hoc baseline (RIO \cite{RIO}), but dedicated probabilistic forward models as well (probPCA \cite{tothova2018, tothova2020}, and MC Dropout \cite{MCDropout}, all of which tended to be overconfident in incorrect predictions. 

The main limitation of Deep MH is its computational complexity. Some of it is because this is a sampling technique. 
The rest stems from the the proposed evaluation of likelihood. 
While tweaks can be done, such as increasing early stopping tolerances in input backpropagation optimisation, a more serious avenue of future exploration involves reformulation of the original optimisation problem via an additional latent variable, which would allow for a more significant relaxation of the full convergence criterion. 
A possible alternative to the proposed likelihood computation via optimisation is the simulation of an inversion process via a generative adversarial network (GAN). However, this is a rather involved solution depending on training of a complex dedicated model potentially requiring large amounts of data.


Deep MH can be applied to problems where it is possible to define or estimate a prior distribution $p(y)$. 
As a result, the effective dimensionality of the target variable $y$ needs to be relatively small so that plausible samples can be generated with ease. This restriction is not very limiting in practice, however. Firstly, problems where the target variables are relatively low dimensional are numerous among current DL applications, e.g., in image recognition. Secondly, as we saw in Sec.~\ref{sec:prior} in some applications, such as shape regression, high dimensional targets can be approximated by low dimensional representations. Applications to problems with higher dimensional targets is an open research question.

In this work, we applied Deep MH to regression problems, however, it can be used for classification tasks without substantial modification, provided $p(y)$ can be evaluated.
This can be possible for example in image-wise classification problems with class probabilities estimated through occurrence numbers in a data set. 

\clearpage

\bibliography{bibliography}

\newpage
\appendix
\onecolumn
\section{Approximate Bayesian Computation}\label{app:ABC}

Approximate Bayesian Computation (ABC) methods \cite{ABC:Rubin} are Bayesian sampling methods that bypass the computation of the likelihood. Instead, data X' is simulated from the likelihood distributions for every proposal $y'\sim p(y)$. The proposals are then only accepted as samples from the posterior $p(y|x)$ if the simulated X' and observed test data set X are \textit{similar enough} as defined by a specific pre-defined distance measure based on a given metric (e.g. squared Euclidean distances), summary statistic (e.g. sample mean) and a threshold level. Due to its parallels to the proposed method presented in Sec.~\ref{sec:MCMC:method}, we include the basic ABC sampler \cite{ABC:pritchard, ABC:review} steps in Algorithm \ref{alg:MCMC:ABC}. This original rejection scheme might suffer from inefficiencies as sampling from the prior leads to proposed values located in low posterior probability regions. Hence, extensions such as MCMC-ABC \cite{MCMC-ABC} incorporating ABC and MCMC, have been proposed.  Note that ABC methods are often referred to as \textit{likelihood-free} \cite{ABC:review}. However, a feasibly formulated likelihood model that can be sampled from is still needed. Hence, when trying to apply ABC to the DL problems, we encounter the same kind of inversion difficulties described in the introduction section. Moreover, it might be costly to simulate multiple samples at once. Note that identification of the right kind of sufficient summary statistics and distance metrics 
is a subject of continued research \cite{ABC:review}. 
\begin{algorithm}[h!]
\caption{ABC rejection sampler \cite{ABC:pritchard}}
\label{alg:MCMC:ABC}
\begin{algorithmic}
\STATE{\textbf{Given} input test data set X $\in \mathcal{X}$ of size $n$, prior on values $y \sim p(y)$, summary statistics $\eta$, distance $\rho$, tolerance level $\epsilon >0$}
 \FOR{t = 1 to L}
    \REPEAT
        \STATE{Generate $y'$ from the prior distribution \py{}}
        \STATE{Generate data set X' of size $n$ from the likelihood $p(\cdot|y')$}
    \UNTIL{the similarity condition $\rho\{\eta(X), \eta(\mbox{X'})\} \leq \epsilon$ is fulfilled}
    \STATE{Set $y_t$ = $y'$.}
 \ENDFOR
\end{algorithmic}
\end{algorithm}

\section{$p(x|y)$ as Compound Distributions}\label{app:compound_dist}
 Example from our experiments: When elaborating (\ref{eq:MCMC:our_likelihood}) for the distance measure $\rho(\cdot, \cdot) = \textrm{MSE}(\cdot,\cdot)$ one arrives at
\begin{equation*}
    p(x|y) \propto \exp \left( -\frac{1}{2}(x-x'_{y'})^{T} (\frac{n}{2\beta} I) (x-x'_{y'})\right)
\end{equation*}
which is not a simple Gaussian around $x'_{y'}$ but rather a compound distribution as $x'_{y'}$ is a random variable drawn from analytically intractable distribution specified by the network. Then even when choosing the target prior as $p(y) = \mathcal{N}(0,I)$ one in general does not arrive at a straightforward Gaussian posterior, as the prior in not conjugate for the compound likelihood.
\section{Experimental Setup}\label{app:experiment_setup}
The left ventricle myocardium boundary delineation testing was performed on the small FOV ($60 \times 60$ pixels) and large FOV ($200 \times 200$ pixels) cardiac MR UK BioBank data sets~\cite{UKBB} described in Sec.~\ref{sec:datasets}. These data sets are imbalanced when it comes to location and orientation of the images with propensity towards  the myocardium being localised in the central area of the image (roughly 90\% of the examples). The original sets were randomly split into test, train and validation subsets. In both settings, the target variable $y \in \mathbb{R}^{50\times2}$ is represented as a contour of 50 connected vertices of fixed connectivity. The original sets were randomly split into test, train and validation subsets.

The proposed uncertainty estimation was tested on a deterministic 10-layer deep surface prediction CNN, analogous to the one proposed in the probPCA paper \cite{tothova2018}, see Fig.~\ref{ap:fig:architecture}.  The network had been trained using data from the training set until full convergence before Deep MH was deployed. The uncertainty methods were calibrated on the validation set. Uncertainty results reported in Sec.~\ref{sec:results} were obtained on data from test sets.

The CT slice localisation task used the same data set (CT data set~\cite{dataset_CT_slice, ML_repository_datasets} described in Sec.~\ref{sec:datasets}) and following the same setup as in the original RIO paper \cite{RIO}. We first trained a fully connected network with 2 hidden layers (256 units in each) for 1000 epochs. Deep MH and RIO ran on the same fixed pre-trained network.

No augmentation was used for training. Computations were done on a GeForce RTX 2080 Ti GPU using PyTorch 1.9. For software legacy reason, RIO and probPCA computations were run on a CPU using TensorFlow 1.12.

Input backpropagation during Deep MH sampling was carried out until convergence determined by early stopping when the gradient of each component of loss (\ref{eq:backprop}) fell under the specified tolerance. Inspired by the original style transfer paper \cite{style_transfer}, we chose the L-BFGS optimiser due to its relatively quick convergence, as examined in \cite{lbfgs_conv, lbfgs_conv2}. 

The MCMC sampling can be time intensive. Fortunately, batching capabilities of current DL frameworks allow for considerable speed ups by means of straightforward chain parallelisation. Parallel runs of independent chains initialised at randomly dispersed locations are also a good tool for minimising auto-correlation within the sample set.

Detailed setup can be found in Tab.~\ref{tab:MCMC:results:setup}.

\begin{figure}[t]
\centering
	\includegraphics[angle=0, origin=c, width=0.7\linewidth]{chapters/MCMC/figures/architecture.pdf}
	\caption{Architecture of the feed-forward network. Column-wise matrix of principal vectors $U$,  principal component diagonal covariance matrix $S$, data mean $\mu$ the were precomputed on the training set. Each convolutional layer (kernel size = 3, stride = 1) was followed by batch normalisation and ReLU activation. Max pooling was done with a kernel of size 2 with a stride of 2. Dropout layers were added before every weight layer for MC Dropout \cite{MCDropout}.}\label{ap:fig:architecture}
\end{figure}

\subsection{Parameter Selection}\label{app:params}

Hyperparameter search was performed on the validation set by hand tuning or, wherever necessary, grid search. The focus of the parameter calibration search were proper optimisation of all the components of the optimisation problem~\ref{eq:backprop}, good convergence times and acceptance rate. While the optimal acceptance rate is a matter of debate and heuristics in the MCMC community~\cite{gelman_handbook}, we strived for an average acceptance of $\sim 20\%$ for the high-dimensional shape predictions tasks and $\sim 40\%$ for the one dimensional CT slice localisation task (the usual recommendation developed on simplified problems states the optimal rates to be 23.4\% and 44\% for high and one-dimensional problems respectively~\cite{MCMC:optimal_acc_rate_highdim, MCMC:optimal_acc_rate_onedim}.

\begin{table*}[t]
    \centering
    \begin{small}
    \begin{tabular}{lccc}
        \hline
        \textbf{Parameter} & \textbf{Myocardium SmallFOV} &\textbf{Myocardium LargeFOV} &\textbf{CT slice localisation} \\
        \hline
           Data set split  & 572|97|100 &  1532|79|299 & 34240|70|7560 \\
            test|train|val & & &\\
           Input data  & MR slice $60\times 60$ & MR slice $200 \times 200$  & slice descriptor $384 \ \times 1$ \\
           Target data & vertex coordinates $50 \times 2$ &  vertex coordinates $50 \times 2$ & relative slice location $1 \times 1$\\
           PCA components & 20 & 20 & - \\
           \# samples & 90000 & 52550 & 1\,mil\\
        \hline
        \multicolumn{4}{ l}{\textbf{Forward network training setup}} \\ \hline
        Epochs & 4210& 7189 & 1000 \\
         &  until full convergence & until full convergence & \\
        Optimiser|LR & Adam|0.0001 & Adam|0.0001 & RMSProp|0.001 \\
        Target loss & MSE & MSE & MSE \\
       \hline
        \multicolumn{4}{ l}{\textbf{Deep MH}} \\ \hline
        Target prior & $\mathcal{N}(0,I)$ & $\mathcal{N}(0,I)$& $\mathcal{U}([0,100])$ \\
        \# chains & 5 & 5& 50 \\
        Burn in & 2000& 2000& 1000\\   
        Proposal  &  $z'\sim\mathcal{N}(z^t, \sigma_z^2 I)$ & same as SmallFOV, &   $y'\sim\mathcal{N}(y^t, I)$\\
        
         &  $s'\sim\mathcal{N}(s^t, \sigma_s^2 I)$ &   $\sigma_s =9$, $\sigma_z = 0.2$ & \\
         &   $\sigma_z = 0.1$, $\sigma_s =2$ & & \\
        $\beta$ & 10000& 30000& 60000\\
        \multicolumn{4}{ c}{\textit{Input backpropagation}} \\
        Optimiser|LR & L-BFGS|1 & L-BFGS|1 & L-BFGS|1 \\
        Early stopping tol & 0.008 & 0.008 & 1e-8\\
        $\lambda$ & 1 & 1 & 1 \\    
        $\sigma_n$ & 0.1 & 0.1 & 0.1\\    
        %
        $\rho, \mu$ & MSE & MSE & MSE \\
        \hline
        \multicolumn{4}{ l}{\textbf{probPCA}} \\ \hline
        $\lambda$ &  1000 & 1000 & -\\
        $\sigma$ & 0.05 & 0.05 &- \\
        \hline
        \multicolumn{4}{ l}{\textbf{MC Dropout}} \\ \hline
        $p$ & 0.5 &  0.5 & 0.3\\
        \hline
        \multicolumn{4}{ l}{\textbf{RIO}} \\ \hline
        Kernel & - & - & RBF \\
        \# inducing points & - & - & 50 \\
        Optimiser  & - & - & L-BFGS-B \\
        Max \# of iterations & - & - & 1000\\
        \hline
        \end{tabular}
    \end{small}
    \caption{Experimental setup.}
    \label{tab:MCMC:results:setup}
\end{table*}

\subsection{Data Sets}\label{sec:datasets}
\subsubsection{UKBB 2D Data Set}
We have tested the proposed method on a task of delineation of myocardium boundaries in cardiac MRI using imaging volumes from UK BioBank \cite{UKBB}. The~2D surface delineation entails prediction of 50 vertices and was evaluated on small ($60 \times 60$ crops around the heart, isotropic pixel of size 1.8\,mm) and large ($200 \times 200$ crops,  isotropic pixel of  size 1.8269\,mm) field of view images.
The short axis 2D cardiac image data set used in our experiments was extracted from full 3D cardiac volumes.  A single slice per volume was extracted at approximately the same vertical position (4th vertical slice) and the same phase of the cardiac cycle (end diastole) across the subjects, resampled to an isotropic pixel size of 1.8 and cropped to the desired size. Training delineations of 50 connected vertices were obtained using active contours \cite{Kass88snakes} from  reference volume segmentations. An automatic method based on expert-segmentations and a combination of learning and registration methods \cite{bai2017, sinclair2017} was used to obtain the segmentations. The connectivity of delinations was assumed constant across the data set and ensured thanks to an automatic detection of the landmark point at the intersection of the right ventricle and left ventricle myocardium. The landmark was selected as the starting point of the oriented contour.
The data set is imbalanced with majority of the samples being roughly centered and aligned by design of MR image acquisition process and human anatomy.

\subsubsection{CT Data Set}

The CT database  \cite{dataset_CT_slice, ML_repository_datasets} consists of a set of descriptors of 2D CT imaging slices of the human head and torso.
It was constructed from 98 3D CT scans (38 neck and  60  thorax  scans) from 74 different patients (43 male, 31 female; age 4--86 years) and covers the continuous area between the top of the head and end of the tailbone (coccyx). Each patient contributed a maximum of 1 neck and 1 thorax scan. The resolution of the CT scans varies between 0.43--0.92 mm in the $x$ and $y$ axes (i.e. across the body), and 66--1,749  slices  per scan (0.5--5 mm/slice) along the z-axes (i.e. along the body). Each slice descriptor comprises two histograms describing the tissue content of the image in the polar space. The first histogram represents the distribution of bone structures, while the second the second the location and distribution of the soft tissue and air. The values of bins outside of the region of interest are set to -0.25. Both descriptors are concatenated into a final feature vector. The target ground truth variable marking the location of the slice within the body was constructed by interpolation between expert manual annotations of distinct landmarks (5 in neck scans, 5 in thorax scans). The locations were rescaled onto interval [0, 100] with 0 denoting the top of the head and 100 the end of the coccyx.

\section{Extended Results}\label{app:results}

In this section we include further visualisations in support of our results presented in Sec.~\ref{sec:results}:
\begin{itemize}
    \item[] \textbf{Fig.~\ref{fig:app:samples}} provides examples of samples from the posteriors presented in Fig.~\ref{fig:samples} in Sec.~\ref{sec:results}. Observe as tightness and variability of the produced samples varies depending on the uncertainty. 
    \item[] \textbf{Fig.~\ref{fig:app:mixing}} visualises convergence behaviour of Deep MH for a subject from Fig.~\ref{fig:mixing} on additional trace plots of the sampled variables. Irrespective of the initialisation, the chains converge to the same equilibrium.
    \item[] \textbf{Fig.~\ref{fig:MCMC:res:acc_confplots}} visualises scaled scatter plots pertinent to confidence plots in Fig.~\ref{fig:MCMC:res:acc_confplots_}. The confidence plots are included here again for ease of comparison. Rescaled scatter plots are useful for diagnosing patterns of behaviours between methods. They allow us to see whether exhibited results are due to the presence of a few outlying values or because of a general trend. 
    \item[] \textbf{Fig.~\ref{fig:app:OOD}} visualises a noise and rotation induced covariate shift $p^{*}_{\textrm{TEST}}(x,y)$ and pertinent Deep MH predictive posteriors. Evaluation of the produced uncertainties is in Sec.~\ref{sec:results}, Fig.~\ref{fig:MCMC:OOD}.

\end{itemize}

%
\begin{figure}[]
\centering
\begin{minipage}[c]{0.7\columnwidth}
    \includegraphics[angle=0, origin=c, width=\linewidth]{chapters/MCMC/figures/MCMC/MCMC_samples.pdf}
\end{minipage}
\caption{50 samples from estimated $p(y|x)$ for subjects from Fig.~\ref{fig:samples}: GT shape (red), forward prediction $f(x_i)$ (blue), sample mean (yellow), individual samples (cyan). Increased variability indicates increased uncertainty.}\label{fig:app:samples}
\end{figure}
\begin{figure}[]
\centering
\begin{minipage}[c]{0.7\columnwidth}
	\includegraphics[angle=0, origin=c, width=\linewidth]{chapters/MCMC/figures/MCMC/mixing/s_trace_cropped.pdf}
\end{minipage}
\\
\begin{minipage}[c]{1\columnwidth}
	\includegraphics[angle=0, origin=c, width=\linewidth]{chapters/MCMC/figures/MCMC/mixing/z_trace_cropped.pdf}
\end{minipage}
\\
\begin{minipage}[c]{1\columnwidth}
	\includegraphics[ angle=0, origin=c, width=\linewidth]{chapters/MCMC/figures/MCMC/mixing/trace_vertex_1024925_cropped.pdf}
\end{minipage}
\\
\begin{minipage}[c]{1\columnwidth}
	\includegraphics[angle=0, origin=c, width=\linewidth]{chapters/MCMC/figures/MCMC/mixing/vertex_trace_dim2_cropped.pdf}
\end{minipage}
\caption{Convergence behaviour of the Deep MH process with 5 chains for subject with poor forward prediction $f(x)$ via trace plots of (top to bottom): $s$ and selected elements of $z$, coordinates of selected vertices, all corresponding to 3rd subject in Fig.~\ref{fig:samples}. Three chains were initialised in the neighbourhood of the GT reference shape, 2 around $f(x)$. Irrespective of the initialisation, the chains converge to the same equilibrium.}\label{fig:app:mixing}
\end{figure}
%

%
\begin{figure*}[]
\captionsetup[minipage]{justification=centering}
\centering
\begin{minipage}{\columnwidth}
\centering
\begin{minipage}[t]{0.4\columnwidth}
	\includegraphics[angle=0, origin=c, height=0.15\textheight] {chapters/MCMC/figures/uncertainty/conf_plots/CT/normed_unc_vs_normed_RMSE_.pdf}
\end{minipage}
\begin{minipage}[t]{0.4\columnwidth}
	\includegraphics[angle=0, origin=c, height=0.15\textheight]
	{chapters/MCMC/figures/uncertainty/conf_plots/CT/acc_vs_confidence_max_.pdf}
\end{minipage}
\caption{CT slice localisation}
\end{minipage}
\\
\begin{minipage}{\columnwidth}
\centering
\begin{minipage}[t]{0.4\columnwidth}
	\includegraphics[angle=0, origin=c, height=0.15\textheight]
	{chapters/MCMC/figures/uncertainty/conf_plots/smallFOV/normed_unc_vs_normed_RMSE_.pdf}
\end{minipage}
\begin{minipage}[t]{0.4\columnwidth}
	\includegraphics[angle=0, origin=c, height=0.15\textheight]
	{chapters/MCMC/figures/uncertainty/conf_plots/smallFOV/acc_vs_confidence_max_.pdf}
\end{minipage}
\end{minipage}
\\
\begin{minipage}{\columnwidth}
\centering
\begin{minipage}[t]{0.4\columnwidth}
	\includegraphics[angle=0, origin=c, height=0.15\textheight]
	{chapters/MCMC/figures/uncertainty/conf_plots/smallFOV/normed_unc_vs_normed_RMSE_ODexRemoved.pdf}
\end{minipage}
\begin{minipage}[t]{0.4\columnwidth}
	\includegraphics[angle=0, origin=c, height=0.15\textheight]
	{chapters/MCMC/figures/uncertainty/conf_plots/smallFOV/acc_vs_confidence_max_ODexRemoved.pdf}
\end{minipage}
\caption{Shape delineation SFOV: full test set (top), homogeneous test subset (bottom).}
\end{minipage}
\\
\begin{minipage}{\columnwidth}
\centering
\begin{minipage}[t]{0.4\columnwidth}
	\includegraphics[angle=0, origin=c, height=0.15\textheight]
	{chapters/MCMC/figures/uncertainty/conf_plots/largeFOV/normed_unc_vs_normed_RMSE_.pdf}
\end{minipage}
\begin{minipage}[t]{0.4\columnwidth}
	\includegraphics[angle=0, origin=c, height=0.15\textheight]
	{chapters/MCMC/figures/uncertainty/conf_plots/largeFOV/acc_vs_confidence_max_.pdf}
\end{minipage}
\end{minipage}
\\
\begin{minipage}{\columnwidth}
\centering
\begin{minipage}[t]{0.4\columnwidth}
	\includegraphics[angle=0, origin=c, height=0.15\textheight]
	{chapters/MCMC/figures/uncertainty/conf_plots/largeFOV/normed_unc_vs_normed_RMSE_ODexRemoved.pdf}
\end{minipage}
\begin{minipage}[t]{0.4\columnwidth}
	\includegraphics[angle=0, origin=c, height=0.15\textheight]
	{chapters/MCMC/figures/uncertainty/conf_plots/largeFOV/acc_vs_confidence_max_ODexRemoved.pdf}
\end{minipage}
\caption{Shape delineation LFOV: full test set (top), homogeneous test subset (bottom).}
\end{minipage}
\caption{Accuracy as a function of uncertainty. Left: Relative uncertainty vs. relative accuracy scatter plot: both axes have been rescaled onto [0,1].  Right: confidence plots of scaled relative uncertainty. All the baselines show the propensity to producing incorrect yet confident predictions.}
\label{fig:MCMC:res:acc_confplots}
\end{figure*}


\begin{figure}[]
\centering
\begin{minipage}[c]{1\columnwidth}
\begin{minipage}[c]{1\columnwidth}
\centering
\begin{minipage}[c]{0.19\columnwidth}
	\includegraphics[angle=0, origin=c, width=\linewidth]{chapters/MCMC/figures/OOD/Noise/level_0.05_MCMC_2123740.pdf}
\end{minipage}
\begin{minipage}[c]{0.19\columnwidth}
	\includegraphics[ angle=0, origin=c, width=\linewidth]{chapters/MCMC/figures/OOD/Noise/level_0.1_MCMC_2123740.pdf}
\end{minipage}
\begin{minipage}[c]{0.19\columnwidth}
	\includegraphics[angle=0, origin=c, width=\linewidth]{chapters/MCMC/figures/OOD/Noise/level_0.2_MCMC_2123740.pdf}
\end{minipage}
\begin{minipage}[c]{0.19\columnwidth}
	\includegraphics[angle=0, origin=c, width=\linewidth]{chapters/MCMC/figures/OOD/Noise/level_0.3_MCMC_2123740.pdf}
\end{minipage}
\begin{minipage}[c]{0.19\columnwidth}
	\includegraphics[angle=0, origin=c, width=\linewidth]{chapters/MCMC/figures/OOD/Noise/level_0.4_MCMC_2123740.pdf}
\end{minipage}
\end{minipage}
\caption{Uncertainty under noise induced domain shift.  Original data corrupted by Gaussian noise $\sim\mathcal{N}(0,\sigma^{2} I)$, $\sigma = 0.05, 0.1, 0.2, 0.3, 0.4$ (left to right). }\label{fig:mcmc:ood:noise:posterior}
\end{minipage}
\\
\begin{minipage}[c]{1\columnwidth}
\centering
\begin{minipage}[c]{1\columnwidth}
\centering
\begin{minipage}[c]{0.138\columnwidth}
	\includegraphics[angle=0, origin=c, width=\linewidth]{chapters/MCMC/figures/OOD/Rotation/angle_-10_MCMC_1047575.pdf}
\end{minipage}
\begin{minipage}[c]{0.138\columnwidth}
	\includegraphics[ angle=0, origin=c, width=\linewidth]{chapters/MCMC/figures/OOD/Rotation/angle_-30_MCMC_1047575.pdf}
\end{minipage}
\begin{minipage}[c]{0.138\columnwidth}
	\includegraphics[angle=0, origin=c, width=\linewidth]{chapters/MCMC/figures/OOD/Rotation/angle_-50_MCMC_1047575.pdf}
\end{minipage}
\begin{minipage}[c]{0.138\columnwidth}
	\includegraphics[angle=0, origin=c, width=\linewidth]{chapters/MCMC/figures/OOD/Rotation/angle_-70_MCMC_1047575.pdf}
\end{minipage}
\begin{minipage}[c]{0.138\columnwidth}
	\includegraphics[angle=0, origin=c, width=\linewidth]{chapters/MCMC/figures/OOD/Rotation/angle_-90_MCMC_1047575.pdf}
\end{minipage}
\begin{minipage}[c]{0.138\columnwidth}
	\includegraphics[angle=0, origin=c, width=\linewidth]{chapters/MCMC/figures/OOD/Rotation/angle_-120_MCMC_1047575.pdf}
\end{minipage}
\begin{minipage}[c]{0.138\columnwidth}
	\includegraphics[angle=0, origin=c, width=\linewidth]{chapters/MCMC/figures/OOD/Rotation/angle_-180_MCMC_1047575.pdf}
\end{minipage}
\end{minipage}
\caption{Uncertainty under rotation induced domain shift. Original data rotated by $\alpha = -10^{\circ}, -30^{\circ}, -50^{\circ}, -70^{\circ}, -90^{\circ}, -120^{\circ}, -180^{\circ}$ (left to right).}\label{fig:MCMC:OOD:rotation}
\end{minipage}
\caption{Visualisation of simulated OOD data and predictive posteriors for increasing levels of degradation. Predictive posterior spread increases as data corruption increases.}\label{fig:app:OOD}
\end{figure}


\section{Uncertainty Estimation Convergence: Are Long Chains Really Needed?}\label{app:convergence}

Depending on the problem size, one of the major drawbacks of (MCMC) sampling methods and hence Deep MH is their computational load. 
This might not be a significant issue in smaller tasks---in Sec. \ref{sec:results} it took roughly 0.15s to produce 50 (parallel) samples in the CT slice localisation task.
The computational cost might be more pronounced when evaluating bigger forward models built for large inputs. In shape delineation for SFOV and LFOV data the average computational cost was more significant: 13s and 23s for 5 parallel proposals respectively.

Since batching capabilities allow as to increase the number of parallel chains at relatively low extra cost\footnote{Proper analysis of speedups enabled by parallelisation is a part of our future work.}, the question is this then: Is it really necessary to let the method run for thousands or tens or thousands steps to reach reliable estimates of uncertainty? While answering this question is in itself its own dedicated area of research with no clear-cut solutions and many proposed heuristics \cite{casella, kruschke1}
, we provide an empirical analysis based on evolution of Deep MH uncertainty estimates over time. 
We need to stress, however, that this analysis can be performed only post-hoc, as running sufficiently long runs is needed to avoid pseudo-convergence \cite{gelman_handbook}.

\begin{figure}[t!]
\centering
\begin{minipage}[c]{0.99\columnwidth}
\centering
\begin{minipage}[c]{0.24\columnwidth}
	\includegraphics[angle=0, origin=c, width=0.93\linewidth]{chapters/MCMC/figures/convergence/CT/new/chainl_vs_uncert_rel_all.pdf}
\end{minipage}
\begin{minipage}[c]{0.24\columnwidth}
	\includegraphics[ angle=0, origin=c, width=\linewidth]{chapters/MCMC/figures/convergence/CT/new/chainl_vs_uncert_abs_all.pdf}
\end{minipage}
\begin{minipage}[c]{0.24\columnwidth}
	\includegraphics[ angle=0, origin=c, width=\linewidth]{chapters/MCMC/figures/convergence/CT/new/chainl_vs_uncertainty.pdf}
\end{minipage}
\begin{minipage}[c]{0.24\columnwidth}
	\includegraphics[ angle=0, origin=c, width=\linewidth]{chapters/MCMC/figures/convergence/CT/new/central_grad_unc_all.pdf}
\end{minipage}
\caption{CT slice localisation. Scale of GT target values: [0,100]}
\end{minipage}
\\
\begin{minipage}[c]{0.99\columnwidth}
\centering
\begin{minipage}[c]{0.24\columnwidth}
	\includegraphics[angle=0, origin=c, width=\linewidth]{chapters/MCMC/figures/convergence/smallFOV/new/chainl_vs_uncert_rel_all.pdf}
\end{minipage}
\begin{minipage}[c]{0.24\columnwidth}
	\includegraphics[ angle=0, origin=c, width=\linewidth]{chapters/MCMC/figures/convergence/smallFOV/new/chainl_vs_uncert_abs_all.pdf}
\end{minipage}
\begin{minipage}[c]{0.24\columnwidth}
	\includegraphics[ angle=0, origin=c, width=0.99\linewidth]{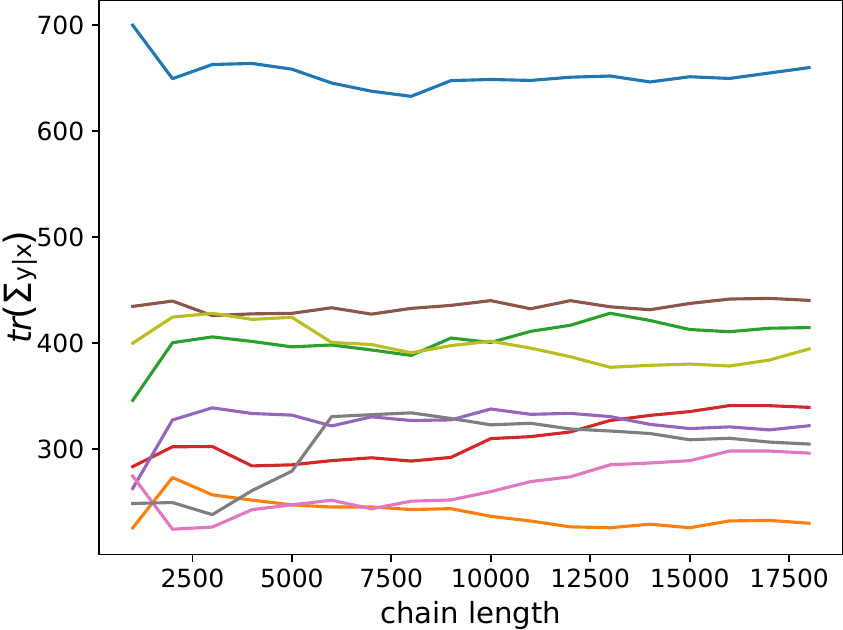}
\end{minipage}
\begin{minipage}[c]{0.24\columnwidth}
	\includegraphics[ angle=0, origin=c, width=\linewidth]{chapters/MCMC/figures/convergence/smallFOV/new/central_grad_unc_all.pdf}
\end{minipage}
\caption{Shape delineation full SFOV. Scale of GT target values: [0,60]}
\end{minipage}
\\
\begin{minipage}[c]{0.99\columnwidth}
\centering
\begin{minipage}[c]{0.24\columnwidth}
	\includegraphics[angle=0, origin=c, width=\linewidth]{chapters/MCMC/figures/convergence/largeFOV/new/chainl_vs_uncert_rel_all.pdf}
\end{minipage}
\begin{minipage}[c]{0.24\columnwidth}
	\includegraphics[ angle=0, origin=c, width=\linewidth]{chapters/MCMC/figures/convergence/largeFOV/new/chainl_vs_uncert_abs_all.pdf}
\end{minipage}
\begin{minipage}[c]{0.24\columnwidth}
	\includegraphics[ angle=0, origin=c, width=\linewidth]{chapters/MCMC/figures/convergence/largeFOV/new/chainl_vs_uncertainty.pdf}
\end{minipage}
\begin{minipage}[c]{0.24\columnwidth}
	\includegraphics[ angle=0, origin=c, width=\linewidth]{chapters/MCMC/figures/convergence/largeFOV/new/central_grad_unc_all.pdf}
\end{minipage}
\caption{Shape delineation full LFOV. Scale of GT target values: [0,200]}
\end{minipage}
\caption{Deep MH uncertainty estimate evolution plots estimated at every 1000 Deep MH steps. The first two columns reflect change over time in comparison to the final value. The third column shows the evolution for randomly selected test subjects with low, high and medium levels of uncertainty. Gradients were  computed using central finite differences at every chain length. Note that in SFOV and LFOV, an absolute change of uncertainty (as measured by the trace of the posterior covariance matrix) of 100 corresponds to an average 1~px change in the standard deviation of a single dimension of the vertex coordinate for an image of size $60\times 60$ and $200 \times 200$ respectively.  
}\label{fig:mcmc:convergence}
\end{figure}

Starting at $t=1000$, we recorded uncertainty estimates every 1000 steps and compared their evolution wirth respect to the values at the maximum chain length for each task as specified in Tab.~\ref{tab:MCMC:results:setup}. Fig.~\ref{fig:mcmc:convergence} shows that across the tasks, the Deep MH estimates tended to be relatively stable from very early on in the sampling process, as evidenced by small relative and absolute errors and near zero gradient values. Taking into account the scale of the target values, this is especially pronounced in the smaller CT localisation tasks, where the estimates undergo negligible change after 1000 Deep MH steps already. 







\section{Role of the Acceptance Rate}\label{sec:MCMC:acc_rate}
While the goal of the hyper-parameter tuning was to maintain an optimal---heuristically recommended---acceptance rate across the data set (an average of $\sim 20\%$ for the low-dimensional and $\sim 40\%$ for the high-dimensional tasks, see Sec.~\ref{app:params}), in practice, it is virtually impossible to guarantee a unified acceptance rate across the board.

Throughout the tasks we observed high correlation between the acceptance rate and uncertainty estimates, see Fig.~\ref{fig:mcmc:acc_rate} (top). In this section we show the acceptance rate is not a confounding variable but rather holds a predictive value when it comes to uncertainty predictions using Deep MH. To do that we adjusted the uncertainty estimates for the effective sample size (ESS). Practically speaking, the chain length (and hence the size of the sample set used to estimate the uncertainty) were adjusted \textit{per subject} so as to maintain constant ESS size across the data set. The correlations were recomputed and are reported in figure \ref{fig:mcmc:acc_rate} (bottom). The chain length adjustment did not lead to significant changes in the uncertainty estimates as compared to full sample set uncertainty estimates, with the average relative changes being -1.2\% (CT), 2.1\% (SFOV), and 0.01\% (LFOV), thus upholding the results reported in the previous sections, see Tab.~\ref{tab:mcmc_acc_rate} and Fig.~\ref{fig:mcmc:acc_rate} for comparison. This indicates that the acceptance rate could be potentially used as an indicator of relative uncertainty within the data set. 
%
%
\begin{table}[h!]
    \centering
    \begin{small}
    \begin{tabular}{rccc}
        \hline
         Pearson Coefficient [r;p] & CT & SFOV & LFOV \\
         \hline
                Without ESS adjustment   & 0.64; 2.03 $\times 10^{-9}$  & 0.62; 1.14 $\times 10^{-11}$  & 0.38; 6.52 $\times 10^{-4}$ \\
                With ESS adjustment   & 0.66; 6.12 $\times 10^{-10}$ & 0.61;  5.53 $\times 10^{-11}$& 0.38; 5.70 $\times 10^{-4}$ \\
         \hline
    \end{tabular}
    \caption{Comparison of results pre and post adjustment for ESS: Pearson coefficients of correlation between uncertainty and accuracy as measured by prediction RMSE.}
    \label{tab:mcmc_acc_rate}
    \end{small}
\end{table}
\begin{figure}[t!]
\centering
\begin{minipage}[c]{0.99\columnwidth}
\centering
\begin{minipage}[c]{0.31\columnwidth}
	\includegraphics[angle=0, origin=c, width=\linewidth]{chapters/MCMC/figures/acc_rate/CT/before/unc_vs_acc_rate.pdf}
	\caption*{\small{\textbf{r = 0.96} \\ p = 0.2 $\times 10^{-39}$}}
\end{minipage}
\begin{minipage}[c]{0.31\columnwidth}
	\includegraphics[ angle=0, origin=c, width=\linewidth]{chapters/MCMC/figures/acc_rate/smallFOV/before/unc_vs_acc_rate.pdf}
	\caption*{\small{\textbf{r = 0.52} \\ p = 0.7 $\times 10^{-9}$}}
\end{minipage}
\begin{minipage}[c]{0.31\columnwidth}
	\includegraphics[ angle=0, origin=c, width=\linewidth]{chapters/MCMC/figures/acc_rate/largeFOV/before/unc_vs_acc_rate.pdf}
	\caption*{\small{\textbf{r = 0.76} \\ p = 0.3 $\times 10^{-16}$}}
\end{minipage}
\end{minipage}
\\
\begin{minipage}[c]{0.99\columnwidth}
\centering
\begin{minipage}[c]{0.31\columnwidth}
	\includegraphics[angle=0, origin=c, width=\linewidth]{chapters/MCMC/figures/acc_rate/CT/after/unc_vs_acc_rate.pdf}
	\caption*{\small{\textbf{r = 0.96} \\ p = 0.2 $\times 10^{-40}$} \\ CT}
\end{minipage}
\begin{minipage}[c]{0.31\columnwidth}
	\includegraphics[ angle=0, origin=c, width=\linewidth]{chapters/MCMC/figures/acc_rate/smallFOV/after/unc_vs_acc_rate.pdf}
		\caption*{\small{\textbf{r = 0.43} \\ p = 0.1 $\times 10^{-5}$} \\SFOV}
\end{minipage}
\begin{minipage}[c]{0.31\columnwidth}
	\includegraphics[ angle=0, origin=c, width=\linewidth]{chapters/MCMC/figures/acc_rate/largeFOV/after/unc_vs_acc_rate.pdf}
		\caption*{\small{\textbf{r = 0.76} \\ p = 0.3 $\times 10^{-40}$} \\LFOV}
\end{minipage}
\end{minipage}
\caption{Correlation between uncertainty and Deep MH and per subject average acceptance rate and pertinent Pearson correlation coefficients. Before (top) and after (bottom) adjustment for ESS.}\label{fig:mcmc:acc_rate}
\end{figure}